\documentclass{article}
\usepackage[nonatbib, final]{include/mlins/mlins}
\usepackage{times}
\usepackage{epsfig}
\usepackage{graphicx}
\usepackage{amsmath}
\usepackage{amssymb}
\usepackage[breaklinks=true,bookmarks=false,colorlinks=true]{hyperref}
\usepackage{multirow}
\usepackage{inconsolata}
\usepackage{subcaption}
\usepackage{bbm}
\usepackage{algorithm}
\usepackage{algorithmicx}
\usepackage{algpseudocode}
\usepackage{program}
\usepackage{listings}
\usepackage{float}
\usepackage{enumitem}
\newcommand{\tsc}[1]{\textsuperscript{#1}}

\hypersetup{
    colorlinks=true,
    urlcolor={cyan},
    linkcolor={blue},
    citecolor={green}
}

\title{Artificial-intelligence-based molecular classification of diffuse gliomas
using rapid, label-free optical imaging}
\author{%
Todd Hollon\tsc{1}\quad
Cheng Jiang\tsc{1}\quad
Asadur Chowdury\tsc{1}\quad
Mustafa Nasir-Moin\tsc{2}\quad\\\textbf{
Akhil Kondepudi\tsc{1}\quad
Alexander Aabedi\tsc{3}\quad
Arjun Adapa\tsc{1}\quad
Wajd Al-Holou\tsc{1}\quad
Jason Heth\tsc{1}\quad}\\\textbf{
Oren Sagher\tsc{1}\quad
Pedro Lowenstein\tsc{1}\quad
Maria Castro\tsc{1}\quad
Lisa Irina Wadiura\tsc{4}\quad}\\\textbf{
Georg Widhalm\tsc{4}\quad
Volker Neuschmelting\tsc{5}\quad
David Reinecke\tsc{5}\quad
Niklas von Spreckelsen\tsc{5}\quad}\\\textbf{
Mitchel Berger\tsc{5}\quad
Shawn Hervey-Jumper\tsc{5}\quad
John Golfinos\tsc{2}\quad
Matija Snuderl\tsc{2}\quad}\\\textbf{
Sandra Camelo-Piragua\tsc{1}\quad
Christian Freudiger\tsc{6}\quad
Honglak Lee\tsc{1}\quad
Daniel Orringer\tsc{2}}\\[1em]
\tsc{1}University of Michigan\quad
\tsc{2}New York University\quad
\tsc{3}University of California San Francisco\quad\\
\tsc{4}Medical University Vienna\quad
\tsc{5}University Hospital Cologne\quad
\tsc{6}Invenio Imaging\\[1em]
\texttt{tocho@med.umich.edu}\qquad
\url{https://deepglioma.mlins.org/}
}
\makeatletter
\def\@noticestring{Preprint by Machine Learning in Neurosurgery Laboratory at the University of Michigan. Manuscript published in \href{https://doi.org/10.1038/s41591-023-02252-4}{\emph{Nature Medicine}}.}
\makeatother

\begin{document}
\maketitle
\begin{abstract}
Molecular classification has transformed the management of brain tumors by enabling more accurate prognostication and personalized treatment. However, timely molecular diagnostic testing for brain tumor patients is limited, complicating surgical and adjuvant treatment and obstructing clinical trial enrollment. Here, we developed DeepGlioma, a rapid ($<$90 seconds), AI-based diagnostic screening system to streamline the molecular diagnosis of diffuse gliomas. DeepGlioma is trained using a multimodal dataset that includes stimulated Raman histology (SRH), a rapid, label-free, non-consumptive, optical imaging method, and large-scale, public genomic data. In a prospective, multicenter, international testing cohort of diffuse glioma patients (N = 153) who underwent real-time SRH imaging, we demonstrate that DeepGlioma can predict the molecular alterations used by the World Health Organization (WHO) to define the adult-type diffuse glioma taxonomy (IDH mutation, 1p19q co-deletion, ATRX  mutation), achieving a mean molecular classification accuracy of 93.3 ($\pm$ 1.6)\%. Our results represent how artificial intelligence and optical histology can be used to provide a rapid and scalable adjunct to wet lab methods for the molecular screening of diffuse glioma patients.

\end{abstract}
\section{Introduction}
Molecular classification is increasingly central to the diagnosis and treatment of human cancers. Diffuse gliomas, the most common and deadly primary brain tumors, are now defined using a handful of molecular markers \cite{Louis2021-vd}. However, molecular subgrouping of diffuse gliomas requires laboratory techniques such as immunohistochemistry, cytogenetic testing and, often, next-generation sequencing that are not uniformly available at the centers where brain tumor patients are treated. Moreover, the expert interpretation of molecular data is increasingly challenging in the setting of a declining pathology workforce \cite{Metter2019-yy}. Consequently, molecular diagnostic and sequencing techniques for brain tumors, when available, is commonly associated with long turnaround times even in well-resourced settings (days-weeks) \cite{Damodaran2015-gq, Fortin_Ensign2021-cr}. Barriers to molecular diagnosis can result in suboptimal care for brain tumor patients complicating prognostic prediction, surgical decision-making, extent of resection goals, selection of adjuvant chemoradiation regimens, and clinical trial enrollment. Here, we propose and prospectively validate an AI-based approach to simplify the molecular classification of diffuse gliomas through automated image analysis of rapid optical imaging of fresh, unprocessed surgical specimens.  

\begin{figure*}[t!]
    \centering
    \includegraphics[width=\textwidth]{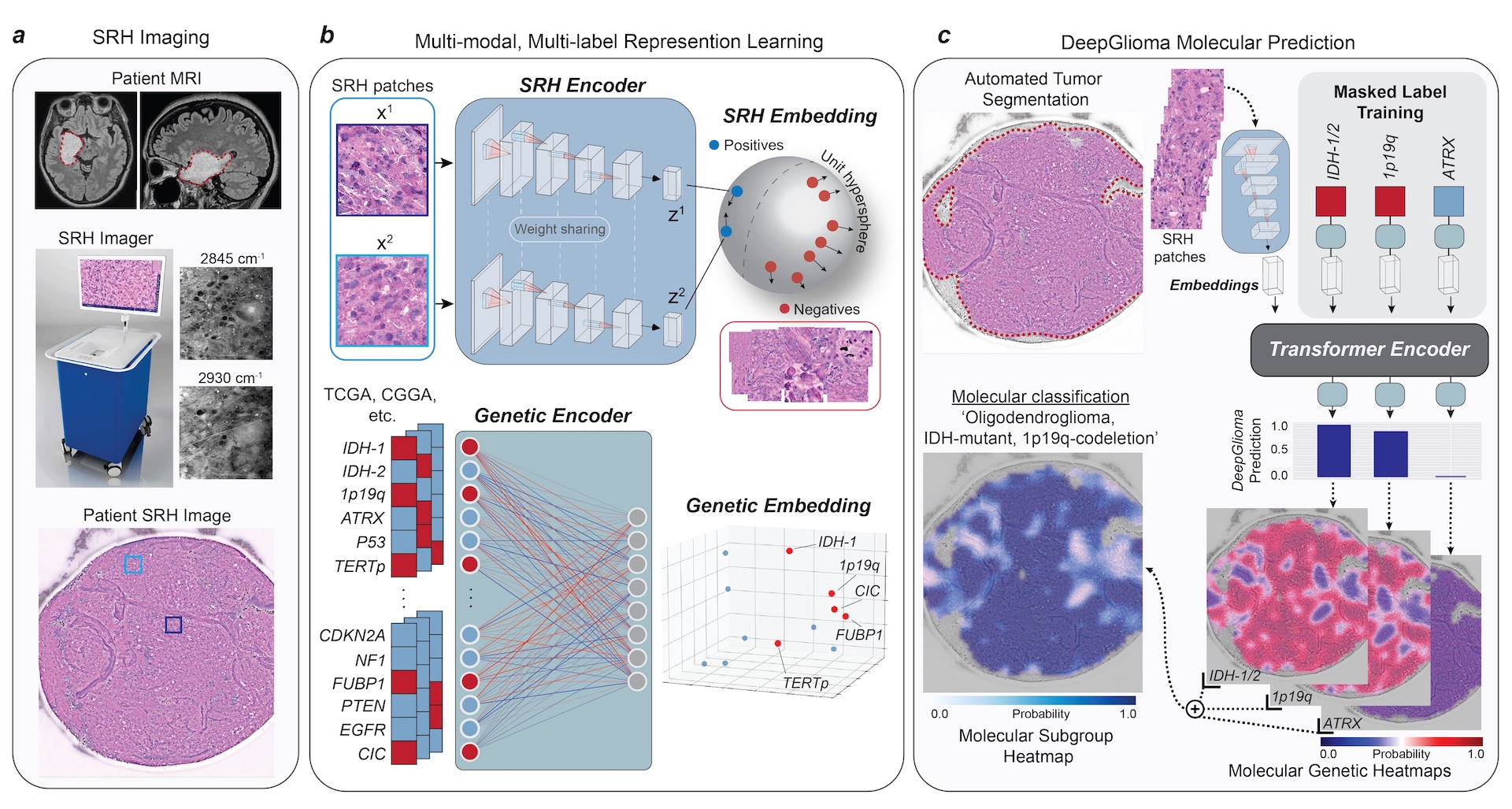}
    \caption{\textbf{Bedside SRH and DeepGlioma workflow}. \textbf{a}, A patient with a suspected diffuse glioma undergoes biopsy or surgical resection. The portable SRH imaging system is used to acquire histologic images in the operating room, performed by a single technician using simple touchscreen instructions. A freshly excised tissue specimen is loaded directly into a custom microscope slide and inserted into the SRH imager without the need for tissue processing (Extended Data Fig. \ref{fig:ex_data1}. Raw SRH images are acquired at two Raman shifts, 2,845 cm\textsuperscript{-1} and 2,930 cm\textsuperscript{-1}, as strips. The time to acquire a 3$\times$3 mm\textsuperscript{2} SRH image is approximately 90 seconds. Raw optical images are rendered using a virtual hematoxlyin and eosin (H\&E)-like lookup table for clinician review \cite{Orringer2017-nn}. \textbf{b}, DeepGlioma is trained using a multi-modal dataset. SRH images are used to train a CNN encoder using weakly supervised, multi-label contrastive learning for image feature embedding (Extended Data Fig. \ref{fig:ex_data3}). Second, public diffuse glioma genomic data from TCGA, CGGA, and others (Extended Data Table 2) are used to train a genetic encoder to learn a genetic embedding (Extended Data Fig. \ref{fig:ex_data5}). \textbf{c}, DeepGlioma molecular prediction is achieved by using a pretrained segmentation model \cite{Hollon2020-ez, Hollon2020-oj} to identify tumor regions, generate patches within those regions, and perform a feedforward pass of tumor patches through the SRH encoder. The SRH and genetic encoders are integrated into a transformer model for multi-label prediction of diffuse glioma molecular diagnostic mutations. To improve DeepGlioma performance, we used masked label training to train the transformer encoder (Extended Data Fig. \ref{fig:ex_data5}). DeepGlioma input is SRH images only during inference. Because our system uses patch-level predictions, spatial heatmaps can be generated for both molecular genetic and molecular subgroup predictions to improve model interpretability, identify regions of variable confidence, and associate SRH image features with DeepGlioma predictions (Extended Data Figs. \ref{fig:ex_data9} and \ref{fig:ex_data10}).}
    \label{fig:workflow}
\end{figure*}

\section{Results}
DeepGlioma is an AI-based diagnostic screening system that combines deep neural networks and stimulated Raman histology (SRH) to achieve rapid molecular screening of fresh glioma specimens (Fig. \ref{fig:workflow}). Our approach predicts the most critical diagnostic genetic alterations in diffuse glioma using learned spectroscopic and histopathologic image features in order to inform patient care and guide downstream definitive molecular testing. Using SRH images only, DeepGlioma can achieve molecular classification in less than 2 minutes of tissue biopsy without the need for tissue processing or human interpretation (Extended Data Fig. \ref{fig:ex_data1}). While DeepGlioma can scale to an arbitrary number of diagnostic mutations, we focus on the major molecular diagnostic alterations used by the WHO CNS5 \cite{Louis2021-vd} and the College of American Pathologists \cite{Brat2022-ve} to define the diffuse glioma subgroups: isocitrate dehydrogenase-1/2 (IDH) mutations, 1p19q chromosome co-deletion (1p19q-codel), and ATRX loss.

The SRH workflow begins when a fresh, unprocessed surgical specimen is biopsied from a brain tumor patient and a small (3$\times$3 mm) sample is placed into a custom microscope slide (Fig. \ref{fig:workflow}a and Extended Data Fig. \ref{fig:ex_data1}). The slide is inserted into the SRH imager and images are acquired at two Raman shifts to generate two image channels: 2,845 cm\textsuperscript{-1} and 2,930 cm\textsuperscript{-1} \cite{Orringer2017-nn}. SRH patches are then sampled in a raster fashion from the whole slide SRH image to generate non-overlapping, single-scale, high-resolution images for model training and inference. We used SRH images from 373 adult diffuse glioma patients treated at the University of Michigan to train a deep convolutional neural network (CNN) as a visual encoder (Extended Data Table 1 and Extended Data Fig. \ref{fig:ex_data2}). Molecular classification is a multi-label classification task, such that the model must predict the mutational status of multiple genetic mutations. While previous studies have used linear classification layers trained end-to-end using cross-entropy \cite{Hollon2020-ez}, we found that weakly supervised (i.e. patient labels only) \underline{patch}-based \underline{con}trastive learning, or \textit{patchcon}, was ideally suited for whole slide SRH classification (Fig. \ref{fig:workflow}b and Extended Data Fig. \ref{fig:ex_data3}) \cite{Chen2020-uz}. We developed a simple and general framework for multi-label contrastive learning of visual representations and trained an SRH encoder using this framework (Extended Data Fig. \ref{fig:ex_data4}). 

Next, we pretrained a genetic embedding model using large-scale, public glioma genomic data (Fig. \ref{fig:workflow}b and Extended Data Table 2). We aimed to learn a genetic embedding space that meaningfully encodes the relationships between mutations in order to improve SRH classification. The co-occurrence of specific mutations in the same tumor defines the molecular subgroups of diffuse gliomas \cite{Eckel-Passow2015-oj, Cancer_Genome_Atlas_Research_Network2015-ms}. The genetic embedding model learns to represent the co-occurrence dataset statistics using global vector embeddings \cite{Pennington2014-dx}. The model learned a linear substructure that matches known molecular subgroups of diffuse gliomas (Extended Data Fig. \ref{fig:ex_data5}). By pretraining an embedding model using a large genomic dataset, DeepGlioma can be trained using the known genomic landscape of diffuse gliomas, allowing for efficient multi-label molecular classification using SRH image features. 

Finally, the pretrained SRH and genetic encoders are integrated into a transformer architecture for multi-label molecular classification (Fig. \ref{fig:workflow}c) \cite{Vaswani2017-in}. During transformer training, the input tokens are the visual embedding of the SRH patch and the genetic embedding for the patient’s tumor. Similar to masked language modeling \cite{Devlin2018-ck}, we randomly mask a subset of the genes from the input and the objective is to predict the masked genes. During inference, the transformer uses only the SRH patch embedding to predict the mutational status of each gene. We performed iterative hold-out cross-validation to show the advantage of patchcon, genetic pretraining, and masked label training through several ablation studies. We demonstrated that DeepGlioma was able to achieve a mean area under the receiver operator characteristic curve (mAUROC) of 92.6 $\pm$ 5.4\% for molecular classification on held-out SRH data. Our multimodal training strategy that included the pretrained genetic embedding model results in $\sim$+5\% increase in overall classification performance (Extended Data Fig. \ref{fig:ex_data6}). 

\begin{figure*}[ht!]
    \centering
    \includegraphics[width=\textwidth]{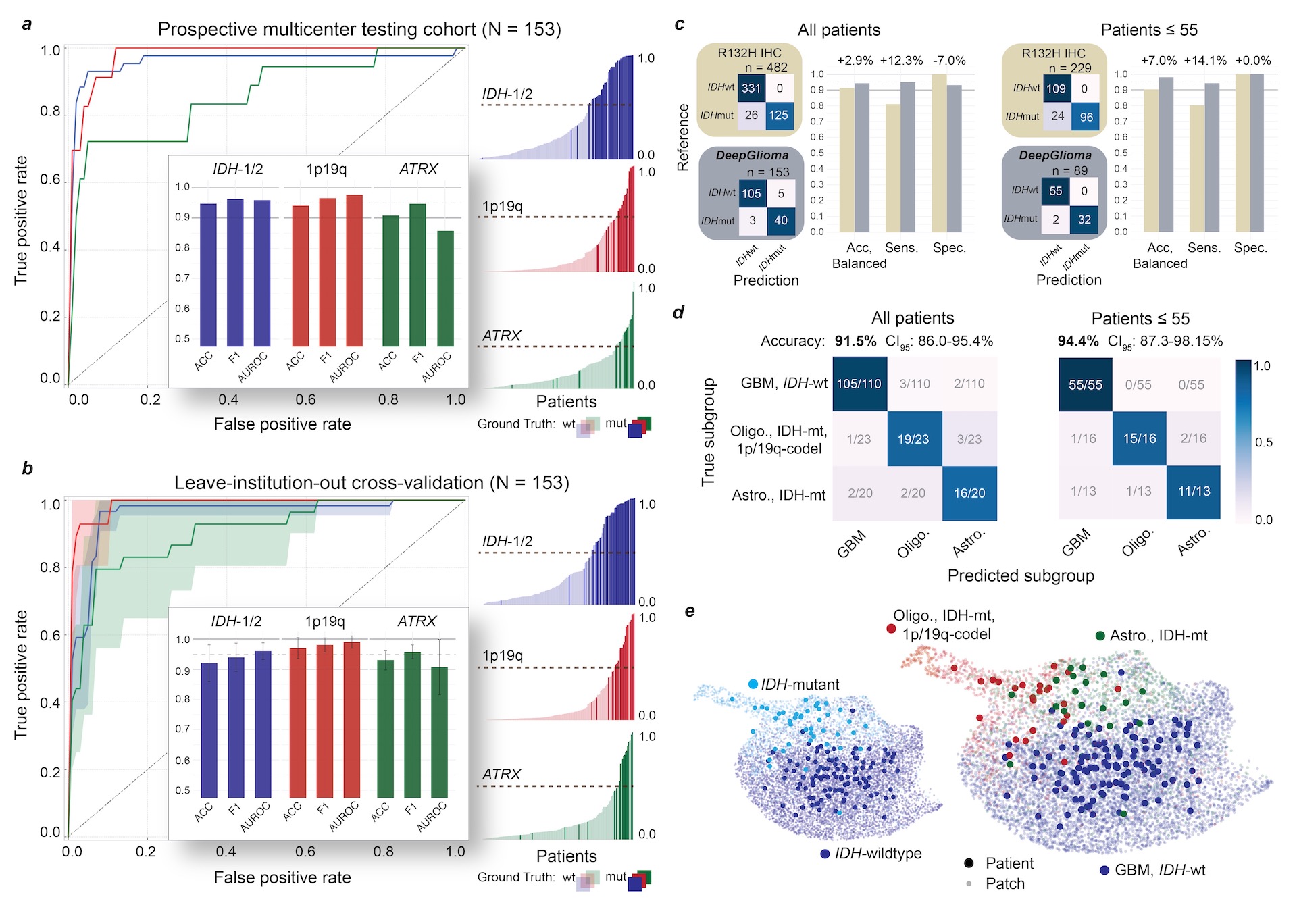}
    \caption{\textbf{DeepGlioma molecular classification performance} \textbf{a}, Results from our prospective multicenter testing cohort of diffuse glioma patients are shown. DeepGlioma was trained using UM data only (n = 373) and tested on our external medical centers (n = 153). All results are presented as patient-level predictions. Individual ROC curves for IDH (AUROC 95.9\%), 1p9q (AUROC 97.7\%), and ATRX (AUROC 85.7\%) classification are shown. Bar plot inset shows the accuracy, F1 score, and AUROC classification metrics for each of the mutations. Individual patient-level molecular genetic prediction probabilities are ordered and displayed. \textbf{b}, Results from the LIOCV experiments. Mean (solid line) and standard deviation (fill color) ROC curves are shown. Metrics are averaged over external testing centers to determine the stability of DeepGlioma classification results given different patient populations, clinical workflows, and SRH imagers. Including additional training data resulted in an increase in DeepGlioma performance, especially for 1p19q and ATRX classification. \textbf{c}, Primary testing endpoint: comparison of IDH1-R132H IHC versus DeepGlioma for IDH mutational status detection. DeepGlioma achieved a 94.2\% balanced accuracy for the prospective cohort and a 97.0\% balanced accuracy  for patients 55 years or younger. The major performance boost was due to the +10\% increase in prediction sensitivity over IDH1-R132H IHC due to DeepGlioma’s detection of both canonical and non-canonical IDH mutations. \textbf{d}, Secondary testing endpoint: DeepGlioma results for molecular subgrouping according to WHO CNS5 adult-type diffuse glioma taxonomy \cite{Louis2021-vd}. Multiclass classification accuracy for all patients and patients 55 years or less are shown. \textbf{e}, UMAP visualization of SRH representations from DeepGlioma. Small, semi-transparent points are SRH patch representations and large, solid points are patient representations (i.e. average patch location) from the prospective clinical cohort. Representations are labeled according to their IDH subgroup and diffuse glioma molecular subgroup. Our patch contrastive learning encourages the SRH encoder to learn representations that are uniformly distributed on the unit hypersphere.}
\label{fig:results}
\end{figure*}

We tested DeepGlioma in a multicenter, prospective cohort of primary, non-recurrent diffuse gliomas to evaluate how our model generalizes across different patient populations, patient care settings, and SRH imaging systems. Model testing was designed as a non-inferiority diagnostic clinical trial. Four tertiary medical centers across the United States and Europe were included as testing recruitment centers. Patients were recruited as a consecutive cohort of adult ($>$18 years) brain tumor patients who underwent biopsy or tumor resection for diffuse glioma. A total of 153 patients were included (Extended Data Table 3). DeepGlioma achieved a molecular classification accuracy for IDH mutation of 94.7\% (95\% CI 90.0-97.7\%), 1p19q-codeletion of 94.1\% (95\% CI 89.1-97.3\%), and ATRX mutation of 91.0\% (95\% CI 85.1-94.9\%), resulting in a mean accuracy of 93.3 $\pm$ 1.6\%. (Fig. \ref{fig:results}a). Despite training and testing dataset imbalance due to different incidences among each mutation, DeepGlioma achieved F1 scores of 96.3\%, 96.6\%, and 94.7\% for IDH, 1p19q codeletion, and ATRX, respectively.

Next, we performed a set of leave-institution-out cross-validation (LIOCV) experiments in order (1) to assess the stability of DeepGlioma performance across medical centers and (2) to determine the effect of increasing training data on model performance (Fig. \ref{fig:results}b). DeepGlioma demonstrated stability across each LIOCV iteration with molecular classification accuracy standard deviation range of $\pm$2.75-6.06\% and a F1 score range of $\pm$1.71-4.70\%. The prediction of ATRX mutations was consistently more challenging across our experiments. We hypothesize that this is related to the heterogeneity of ATRX-mutated tumor cells within IDH-mutant astrocytomas and that ATRX mutations can occur in both IDH-wildtype and mutant gliomas \cite{Cancer_Genome_Atlas_Research_Network2015-ms}. However, our LIOCV results indicate that this challenge can be addressed with additional training data. DeepGlioma LIOCV classification performance of ATRX mutation improved by a minimum of +2\% across all evaluation metrics compared to our prospective clinical testing results.

We compared the performance of DeepGlioma versus the current gold-standard molecular screening modality for diffuse glioma classification: IDH1-R132H immunohistochemistry (IHC). Given that non-canonical (non-R132H) IDH mutations occur in 20-30\% of IDH-mutant lower grade gliomas \cite{DeWitt2017-ju}, IDH1-R132H IHC has known limitations in clinical practice. Due to the higher rates of lower grade gliomas in young patients, genetic sequencing of IDH is recommended for glioma patients 55 years or younger \cite{Louis2017-zj}. Agnostic to IDH isoform, DeepGlioma generalizes to both canonical and non-canonical IDH mutations. IDH1-R132H IHC has a balanced accuracy of 91.4\% (sensitivity 82.8\%, specificity 100\%). In our prospective, multicenter, testing cohort, DeepGlioma achieved a balanced accuracy of 94.2\% (sensitivity 95.5\%, specificity 93.0\%). In patients 55 years or less, IDH1-R132H has a balanced accuracy of 90.0\% and DeepGlioma achieved a balanced accuracy of 97.0\% (Fig. \ref{fig:results}c). Full patient demographic subgroup analyses can be found in Extended Data Fig. \ref{fig:ex_data7} and \ref{fig:ex_data8}. All non-canonical mutations in our prospective cohort were correctly classified by DeepGlioma (Extended Data Fig. \ref{fig:ex_data7}). DeepGlioma's increased sensitivity allows for better screening of young adults for IDH mutations in which non-canonical mutations are more common.

Finally, DeepGlioma’s prediction of the molecular genetics of diffuse gliomas enables direct classification of SRH images into 3 mutually exclusive adult-type diffuse glioma molecular subgroups as defined by the WHO CNS5 classification scheme (IDH-wildtype, IDH-mutant and 1p19q-codel, IDH-mutant and 1p19q-intact) \cite{Louis2021-vd}. An algorithmic inference method was developed to classify each patient into a molecular subgroup (Algorithm \ref{algo1}). We established an AI-based performance benchmark motivated by our previous methods of SRH classification using a ResNet50 model trained for multiclass classification \cite{Hollon2020-ez, Hollon2020-oj}. DeepGlioma achieved a diffuse glioma molecular subgroup classification accuracy of 91.5\% (95\% CI 86.0-95.4\%) (Fig. \ref{fig:results}d) and demonstrated a +4.6\% performance increase over our benchmark model (Extended Data Fig. \ref{fig:ex_data8} and Extended Data Table 4). The major performance gains of DeepGlioma are due to increased sensitivity for identifying IDH-mutant diffuse gliomas and explicitly modeling the co-occurrences of mutations within molecular subgroups. In patients 55 years or younger, our classification performance showed an overall increase (+2.9\%), obtaining a classification accuracy of 94.4\% (95\% CI 87.3-98.2\%) (Fig. \ref{fig:results}d and Extended Data Fig. \ref{fig:ex_data8}). DeepGlioma performance generalized well across multiple medical centers despite distinct patient populations, clinical presentations, personnel, and infrastructure. Molecular subgroup prediction heatmaps for both canonical (see Extended Data Fig. \ref{fig:ex_data9}) and non-canonical IDH mutations (see Extended Data Fig. \ref{fig:ex_data10}) were generated to improve model interpretability and map DeepGlioma predictions to SRH image features. High-resolution molecular genetic and molecular subgroup predictions can be accessed through our interactive DeepGlioma website (\url{https://deepglioma.mlins.org}).

\section{Discussion}
We present DeepGlioma, a deep learning-based screening system designed to streamline the detection of key molecular alterations in human gliomas. Here, we show that using only SRH images as input, DeepGlioma can predict the genetic mutations that define the WHO classification of adult-type diffuse gliomas within minutes of tumor biopsy. DeepGlioma accurately predicted IDH mutations, 1p19q-codeletion, and ATRX mutations without the need for fluorescence in-situ hybridization or genetic sequencing, enabling automated molecular subtyping of diffuse gliomas according to the WHO classification scheme \cite{Louis2021-vd}.

Access to molecular diagnostic testing is uneven for patients who receive brain tumor care. DeepGlioma can streamline molecular testing by providing rapid molecular screening, enabling clinicians to focus on confirming the most likely diagnostic mutations only, rather than using a diagnostic shotgun approach \cite{Capper2018-al}. In addition, SRH is not consumptive and does not diminish diagnostic yield of tumor specimens, preserving scant clinical samples for definitive molecular testing.

Streamlining molecular classification could also have an immediate impact on the surgical care of brain tumor patients. A growing body of evidence supports that surgical goals should be tailored based on molecular subgroups \cite{Drexler2022-qy, Molinaro2020-qs}. Molecular astrocytoma patients who undergo gross total resection achieve a 5-year increase in median survival compared to patients who receive subtotal resections ($\sim$12 years versus $>$17 year median survival). DeepGlioma creates an avenue for accurate and timely differentiation of diffuse glioma subgroups in a manner that can be used to define surgical goals with a better-calibrated risk-benefit analysis.

Even with optimal standard-of-care treatment, diffuse glioma patients face limited treatment options. Consequently, the development of novel therapies through clinical trials is essential. Unfortunately, fewer than 10\% of glioma patients are enrolled in clinical trials \cite{Vanderbeek2018-ba}. Clinical trials limit inclusion criteria to a specific sub-population, often defined by genetic subgroups.  By providing an avenue for rapid molecular screening, DeepGlioma can initiate the process for trial enrollment at the earliest stages of patient care. Moreover, DeepGlioma can facilitate clinical trials that rely on intraoperative local delivery of agents into the surgical cavity and circumvent the blood-brain barrier, a major challenge in therapeutic delivery.

Limitations of our study include that the external testing cohort was restricted to the US and Europe, potentially overfitting to this patient demographic. While our subgroup analysis did not show a difference in performance across minority populations, DeepGlioma validation using a diverse, global demographic would improve model testing. Similar to other deep neural networks, DeepGlioma is not directly interpretable. Uncovering the learned optical image features that predict molecular subgroups is an open question for future investigations.

In conclusion, our study demonstrates how AI-based screening methods have the potential to augment existing conventional diagnostic techniques to improve the access and speed of molecular diagnosis and improve the care of brain tumor patients.

\clearpage
\section{Methods}

\subsection{Study design}

The main objectives of the study were to (1) develop a rapid molecular diagnostic screening tool for classifying adult-type diffuse gliomas into the taxonomy defined by the WHO CNS5 \cite{Louis2021-vd} using clinical SRH and deep learning-based computer vision methods and (2) test our molecular diagnostic screening tool in a large multicenter prospective clinical testing set. We aimed to demonstrate that key molecular diagnostic mutations produce learnable spectroscopic, cytologic, and histoarchitectural changes in SRH images that allow for accurate molecular classification. We aimed to make a clinical contribution by demonstrating that our trained diagnostic system, DeepGlioma, could robustly and reproducibly screen fresh diffuse glioma specimens for specific mutations to inform intraoperative decision making and potentially improve early clinical trial enrollment. DeepGlioma consists of two pretrained separable modules, a visual encoder and a genetic encoder, that are integrated using a multi-headed attention mechanism for image classification \cite{Vaswani2017-in}. Inspired by previous work on deep visual-semantic embedding \cite{Frome2013-pd} and text-to-image generation \cite{Ramesh2021-jw, Saharia2022-kp, Radford2021-kp}, our aim was to use multimodal data that included both imaging and genomic data to achieve optimal performance on a multi-label genetic classification task. The primary SRH dataset for model training and validation was from the University of Michigan and the prospective testing dataset was collected from four international institutions: (1) New York University (NYU), (2) University of California, San Francisco (UCSF), (3) Medical University of Vienna (MUV), and (4) University Hospital Cologne (UKK). We focused on predicting the most clinically important molecular aberrations in diffuse gliomas, but aimed to develop a model architecture that could scale to any number of recurrent mutations in human cancers. For the purposes of this study, we focused our classification task on three key molecular aberrations found in adult-type diffuse gliomas: IDH mutation, 1p19q-codeletion, and ATRX mutation.

\subsection{Stimulated Raman histology}
The operating surgeon was instructed to provide a grossly lesional-appearing, but viable tumor for SRH imaging. This strategy applies to all brain tumor biopsies to maximize the chance of sampling diagnostic tissue. Fiber-laser-based stimulated Raman scattering microscopy was used to acquire all images in our study  \cite{Freudiger2008-gj, Freudiger2014-at}. Detailed description of laser configuration has been previously described \cite{Orringer2017-nn}. In brief, surgical specimens were stimulated with a pump beam at 790 nm and Stokes beam that has a tunable range from 1015 nm-f1050 nm. These laser settings allow for access to the Raman shift spectral range between 2800 cm\textsuperscript{-1} - 3130 cm\textsuperscript{-1}. Images were acquired as 1000 pixel strips with an imaging speed of 0.4 Mpixel(s) per strip. We acquire two image channels sequentially at 2845 cm\textsuperscript{-1} (CH2 channel) and 2930 cm\textsuperscript{-1} (CH3 channel) Raman wavenumber shifts. A strong stimulated Raman signal at 2845 cm\textsuperscript{-1} corresponds to the CH2 symmetric stretching mode of lipid-rich structures, such as myelin. A Raman peak at 2930 cm\textsuperscript{-1} highlights protein- and nucleic acid-rich regions such as the cell nucleus. The first and last 50 pixels on the long axis of each strip are removed to improve edge alignment and the strips are concatenated along the long dimension to generate a stimulated Raman histology image \cite{Orringer2017-nn}. A virtual hematoxylin and eosin (H{\&}E) colorscheme can be applied to the two Raman channels to generate a three-channel, virtually-stained RGB SRH image. These images provide a major advantage over conventional H\&E histology because they allow for real-time pathologic review without degradation of diagnostic accuracy. Multiple studies have demonstrated near-perfect diagnostic concordance with 10X time savings \cite{Orringer2017-nn, Hollon2018-hl}.  These images are used for clinician interpretation and designed to replicate the image contrast seen in conventional H{\&}E histology, but are not used for model training. An overview of SRH imaging workflow can be found in Extended Data Fig. \ref{fig:ex_data1}

\subsection{Image data processing}
All model training and inference was done using the raw, non-virtually colored SRH images. All images were acquired, processed, and archived as 16-bit images to retain spectroscopic image features. Each strip has a 900-pixel width (i.e. after edge clipping) and up to 6000-pixel height. Field flattening correction is used to correct for variation in pixel intensities within image strips. To account for tissue shifts that occur during and between image channel acquisition, the sequentially acquired CH2 and CH3 strips are co-registered using a discrete Fourier transform-based technique for translation, rotation, and scale-invariant image registration \cite{Reddy1996-zl}. Following registration, a pixel-wise subtraction between the CH3 and CH2 channels generates a third ‘red’ channel that highlights the cell nuclei and other protein-rich structures. The whole slide SRH images are finally split into 300x300-pixel patches without overlap using a sliding raster window over the full image. SRH patches are then classified into one of three classes, tumor, normal brain, or nondiagnostic tissue, using our previous trained whole slide SRH segmentation model \cite{Hollon2020-ez, Hollon2020-oj}. Only tumor regions are used for DeepGlioma training and inference (Extended Data Fig. \ref{fig:ex_data1}).

\subsection{Patient enrollment and training dataset generation}
Clinical SRH imaging began at the University of Michigan on 1 June 2015 following Institutional Review Board approval (HUM00083059). Our imaging dataset was generated using two SRH imaging systems. An initial prototype SRH imager \cite{Orringer2017-nn} and the NIO Imaging System (Invenio Imaging, Inc., Santa Clara, CA) \cite{Hollon2020-ez}. All patients with a suspected brain tumor are approached for intraoperative SRH imaging. Inclusion criteria for SRH imaging were patients who were undergoing surgery for (1) suspected central nervous system tumor and/or (2) epilepsy, (3) subject or durable power of attorney was able to provide consent, and (4) preoperative expectation that additional tumor tissue would be available beyond what is required for clinical pathologic diagnosis. Exclusion criteria were (1) insufficient diagnostic tissue as determined by surgeon or pathologist, (2) grossly inadequate tissue (e.g. hemorrhagic, necrotic, fibrous, liquid, etc.), or (3) SRH imager malfunction. Following intraoperative SRH imaging, inclusion criteria for the diffuse glioma training dataset were the following: (1) 18 years or older and (2) final pathologic diagnosis of an adult-type diffuse glioma as defined by WHO CNS5 classification \cite{Louis2021-vd}. Exclusion criterion was less than 10\% area segmented as tumor by our trained SRH segmentation model. UM dataset generation was stopped on 11 November 2021 and a total of 373 patients were included for model training and validation. Patient demographics and molecular diagnostic information can be found in Extended Data Table 1 and Extended Data Fig. \ref{fig:ex_data2}.

\subsection{Multi-label contrastive visual representation learning}
Visual representation learning entails learning a parameterized mapping from an input image to a feature vector that effectively represents the most important image features for a given computer vision task. We used a ResNet50 architecture \cite{He2016-tr} for SRH feature extraction and did not find that larger models provided better performance. While much of our previous work used conventional cross-entropy loss functions to train deep neural networks \cite{Hollon2020-ez, Hollon2020-oj, Orringer2017-nn}, we found that contrastive loss functions result in better visual representation learning \cite{Jiang2022-ea, Chen2020-uz}. We trained our model using a supervised contrastive loss:
\begin{equation}
    \mathcal{L}^\text{sup} = \sum_{i \in I} \mathcal{L}^\text{sup}_i = -\sum_{i \in I} 
    \frac{1}{\vert P(i) \vert}\\ \sum_{p \in P(i)}
    \log \frac {\exp(\text{sim}(g(\boldsymbol{z}_i), g(\boldsymbol{z}_{p}))/\tau)} {\sum_{n \in A(i)} \exp(\text{sim}(g(\boldsymbol{z}_i), g(\boldsymbol{z}_n)) /\tau)}
\end{equation}
where $\boldsymbol{z}=f(\boldsymbol{x}) \in \mathbb{R}^d$ is the d-dimensional feature vector of image x after a feedforward pass through the visual encoder $f(\cdot)$. A linear projection layer $g(\cdot)$ maps the image feature vector $\boldsymbol{z}_i$  to a 128-dimensional space where the contrastive objective is computed. $\boldsymbol{z}_p$ is a feature vector from the set of paired positive examples, $P(i)$, for feature vector $\boldsymbol{z}_i$, and $A(i)$ is the set of all images in a minibatch. $\tau \in \mathbb{R}^{+}$ is a temperature hyperparameter. Paired positive examples are images sampled from the same label. The cosine similarity metric was used in the contrastive objective function, $\mathrm{sim}(\boldsymbol{u},\boldsymbol{v})= \frac{\boldsymbol{u} \cdot \boldsymbol{v}}{\lVert\boldsymbol{u}\rVert\lVert\boldsymbol{v}\rVert}$, to enforce that all feature vectors are on the unit hypersphere. We developed a novel framework for supervised contrastive learning to accommodate multi-label classification tasks. Each label is assigned a unique projection layer $g_\ell(\cdot)$ for computing a label-wise supervised contrastive objective. The final weighted multi-label supervised contrastive loss is:
\begin{equation}
    \mathcal{L}^\text{sup}_\text{multi-label} = \sum_{\ell \in L} \lambda_\ell \sum_{i \in I} \mathcal{L}^\text{sup}_i(i, g_\ell(\cdot), P_\ell(i))
\end{equation}
where $\mathbf{\lambda}_\ell$ is the label weight coefficient. The PyTorch-style pseudocode for implementation can be found in Extended Data Fig. \ref{fig:ex_data3}. All models were trained for 50 epochs using the Adam optimizer with an initial learning rate of 0.001, a cosine annealing learning rate scheduler, and a temperature of 0.07. The batch size was 256. Data augmentation included random cropping, gaussian blur, flipping, and random erasing. Following training, all projection layers were discarded and the visual encoder $f(\cdot)$ was retained for multi-label classification training. We call the above visual representation learning strategy patchcon for weakly supervised (e.g. patient labels only), \underline{patch}-based \underline{con}trastive representation learning, and results can be found in Extended Data Fig. \ref{fig:ex_data4}.

\subsection{Diffuse glioma genetic embedding}
A major component of our multimodal training method includes public genomic data from adult diffuse glioma patients to pretrain a genetic embedding model. We aggregated genomic data from The Cancer Genome Atlas (TCGA), Chinese Glioma Genome Atlas (CGGA) \cite{Zhao2021-gy}, International Cancer Genome Consortium (ICGC) \cite{Zhang2019-ag}, Rembrandt brain cancer dataset \cite{Gusev2018-xt}, Memorial Sloan Kettering (MSK) Data Catalog \cite{Jonsson2019-eu}, and Mayo Glioblastoma Xenograft National Resource. A total of 2777 patients with diffuse gliomas were aggregated and used for embedding model training. The data used to train our genetic embedding model can be found in Extended Data Table 2. Briefly, we selected common recurrent somatic mutations found in adult-type diffuse gliomas and encoded those mutations as either mutant or wildtype for each patient. Inspired by previous work on word embeddings \cite{Pennington2014-dx}, we used a global vector (GloVe) embedding loss function that minimizes the mean squared difference between the pairwise inner product of the learned gene embedding vectors and the co-occurrence of the genes mutational status.
\begin{equation}
\mathcal{L}_\text{embed} = \sum_{i,j} f(\boldsymbol{X}_{i,j}) (\boldsymbol{e}_i^{\top}\boldsymbol{e}_j - \log \boldsymbol{X}_{i,j})^2
\end{equation}
$\boldsymbol{X} \in \mathbb{R}^{2n \times 2n}$ is the pairwise gene co-occurrence matrix for our dataset,  where $\boldsymbol{X}_{i,j}$ is the number of times the mutational status of the $i$-th and $j$-th genes co-occurred in the same tumor. $n$ is the number of genes. The vectors $\boldsymbol{e}_i$ and $\boldsymbol{e}_j$ are updated to match the gene co-occurrence in our dataset. $f(\cdot)$ is a weighting function as previously described to avoid overweighting the most common co-occurrence pairs \cite{Pennington2014-dx}. We found that global vector embeddings perform better than Gene2Vec embedding models \cite{Du2019-yw}. The embedding model is trained for 10K epochs with a batch size of 60. The Adam optimizer was used with a learning rate of 5E-5. Pretrained genetic embedding results can be found Extended Data Fig. \ref{fig:ex_data5}. This method of using multimodal datasets can be extended to other clinical or imaging modalities, such as patient demographics or preoperative/intraoperative magnetic resonance imaging.

\subsection{Multi-label molecular classification}
Two multi-label molecular classification strategies were tested, a linear binary relevance strategy and a transformer-based strategy. Linear binary relevance involves splitting a multi-label classification task into multiple independent binary classifiers. The advantage of using a transformer-based strategy for multi-label classification is the ability to explicitly model complex label dependencies and the co-occurrence of specific genetic mutations in the context of pretrained visual features using an attention mechanism. Similar to bidirectional masked language modelling in BERT-style pretraining \cite{Devlin2018-ck}, we randomly mask a subset of the genetic mutations from the input and the objective is to predict the unknown or masked genes. Masked label training allows for more semantically informative supervision during model training that can improve multi-label classification performance.

\paragraph{Linear binary relevance strategy} Following the training of our visual encoder $f(\cdot)$ using supervised contrastive learning, the weights are fixed and a multilayer perceptron (MLP) that contains a single linear layer is added and trained for multi-label classification. 
\begin{equation}
\hat{\boldsymbol{y}}_\ell = \text{MLP}_\ell(f(\boldsymbol{x})) = \sigma((\boldsymbol{W}_\ell \cdot f(\boldsymbol{x})) + \boldsymbol{b}_\ell)
\end{equation}
where $\sigma$ is a sigmoid activation function that outputs the probability for the $\ell$\textsuperscript{th} genetic mutation. This layer is trained using a weighted binary cross-entropy loss:
\begin{equation}
    \mathcal{L}(\boldsymbol{y}, \hat{\boldsymbol{y}}) = \sum_{\ell = 1}^{\vert L \vert} \lambda_\ell [\boldsymbol{y}_\ell\log(\hat{\boldsymbol{y}}_\ell) + (1-\boldsymbol{y}_\ell) \log(1-\hat{\boldsymbol{y}}_\ell))
\end{equation}

\paragraph{Transformer-based strategy} A transformer encoder is used that includes our pretrained genetic embedding layer $\boldsymbol{W}_\ell$. The labels $[\boldsymbol{\ell}_j, \dots, \boldsymbol{\ell}_k]$ are embedded such that $\boldsymbol{e}_k = \boldsymbol{W}_\ell \cdot \boldsymbol{\ell}_k$ where the $k$\textsuperscript{th} column of $\boldsymbol{W}_\ell$ is the label embedding for the $k$\textsuperscript{th} label. A label mask is then sampled that randomly selects a subset of labels for transformer input and the remainder to be predicted as output. We used learnable state embeddings to encode whether a label was positive, negative, or unknown/masked (not included to simplify notation) \cite{Lanchantin2020-mz}. The image feature vector $\boldsymbol{z}$ and embedded genetic labels are concatenated and input into the transformer encoder: 
\begin{equation}
\boldsymbol{H} = \text{MultiHeadAttention}([\boldsymbol{z}, \boldsymbol{e}_j, \dots, \boldsymbol{e}_k])
= [\boldsymbol{h}_j, \dots, \boldsymbol{h}_k] 
\end{equation}
where $\boldsymbol{H} = [\boldsymbol{h}_j, \dots, \boldsymbol{h}_k]$ are the output representations of the genetic labels and the image token removed. Rather than using a position-wise linear feedforward network and/or a [class] token for label classification as is does in conventional transformer architectures \cite{Lanchantin2020-mz, Vaswani2017-in, Dosovitskiy2020-xs}, we enforce that the output latent space of the transformer encoder is the same as the pretrained genetic embedding space such that:
\begin{equation}
\hat{\boldsymbol{y}} = \sigma(\text{diagonal}(\boldsymbol{H}\boldsymbol{W}_\ell^{\top}))
\end{equation}

where $\boldsymbol{H}\boldsymbol{W}_\ell^{\top}$ is in $\mathbb{R}^{\ell \times \ell}$ matrix and the diagonal elements are the inner product between transformer output latents and the corresponding label embedding of the same label index. The transformer encoder model is trained using the same weighted binary cross-entropy loss function as above. The embedding layer weights are fixed during the transformer encoder training. The PyTorch-style pseudocode for implementation can be found in Extended Data Fig. \ref{fig:ex_data5}.

\subsection{Whole slide segmentation, patient inference, and molecular subgrouping}
Patch-based image classification requires an inference function to aggregate patch-level predictions into a single whole slide-level or patient-level diagnosis. To accomplish this, whole slide SRH images are patched and each patch undergoes an initial feedforward pass through our previously trained segmentation model, $f_{\phi}$, that classifies each patch into tumor, normal brain, or nondiagnostic tissue using an argmax operation \cite{Hollon2020-ez}. If less than 10\% of the image area is classified as tumor, the whole slide is excluded from inference for molecular classification. Our DeepGlioma model, $g_{\theta}$, predicts on the tumor patches only. The patch-level model outputs are summed using soft probability density aggregation, and each label is then renormalized to give a valid Bernoulli distribution for each label. For patients with multiple whole slide images, all patch-level predictions are aggregated and a single patient-level diagnosis is returned. The molecular genetic patient inference function is:
\begin{equation}
p^\text{patient}(\boldsymbol{y} \vert \mathcal{X}) = \frac{1}{Z} \sum_{j=1}^{\vert \mathcal{X} \vert}
\mathbbm{1}(\mathrm{argmax}\ p(\boldsymbol{y}_j \vert \boldsymbol{x}_j, \phi) = k_{\text{tumor}}) p(\boldsymbol{y}_j \vert \boldsymbol{x}_j, \theta)
\end{equation}
where $\mathcal{X}$ is a set of patches from a patient, $\boldsymbol{x}_j$  is the $j^{th}$ patch, $p(\boldsymbol{y}_j \vert \boldsymbol{x}_j, \phi)$ is the patch output from the tumor segmentation model $f_{\phi}$, $p(\boldsymbol{y}_j \vert \boldsymbol{x}_j, \theta)$ is the DeepGlioma $g_{\theta}$ output, and $Z = \sum_{j=1}^{\vert \mathcal{X} \vert} \mathbbm{1}(\text{argmax} \: p(\boldsymbol{y}_j \vert \boldsymbol{x}_j, \phi) = k_{\text{tumor}}$ is the number of patches classified as tumor. Mutually-exclusive molecular subgroup prediction is achieved algorithmically from the above patient-level molecular genetic predictions $p^\text{patient}(\boldsymbol{y} \vert \mathcal{X})$ as shown in Algorithm \ref{algo1}.

\begin{algorithm*}
\caption{DeepGlioma patient-level molecular subgroup prediction}\label{algo1}
\begin{algorithmic}[1]
\Require $ p(\boldsymbol{y} \vert \mathcal{X}), \tau, \psi, \epsilon $ \Comment{$\tau = 0.5, \psi = 1$ for DeepGlioma experiments}
\If{$p(y[k_\text{IDH}] \vert \mathcal{X}) < \tau $}\\
    \qquad\Return  ``Glioblastoma, IDH-wildtype"
\ElsIf {$p(y[k_\text{IDH}] \vert \mathcal{X}) \ge \tau \And \displaystyle\frac{p(y[k_\text{1p19q}] \vert \mathcal{X})}{p(y[k_\text{ATRX}] \vert \mathcal{X}) + \epsilon} > \psi$}\\
    \qquad\Return  ``Oligodendroglioma, IDH-mutant, and 1p19q-codeleted"
\ElsIf{$p(y[k_\text{IDH}] \vert \mathcal{X}) \ge \tau \And   \displaystyle\frac{p(y[k_\text{1p19q}] \vert \mathcal{X})}{p(y[k_\text{ATRX}] \vert \mathcal{X}) + \epsilon} \leq \psi$}\\
    \qquad\Return  ``Astrocytoma, IDH-mutant"
\EndIf
\end{algorithmic}
\end{algorithm*}

\subsection{Ablation studies}
We conducted three main ablation experiments to test the importance of major training strategies and model architectural design choices: (1) cross-entropy versus contrastive learning for visual representation learning, (2) linear versus transformer-based multi-label classification, and (3) fully-supervised versus masked label training. Using the UM dataset only, we performed hold-out validation on three randomly sampled validation sets (n = 20 patients/set) that contained a balanced number of IDH mutant (n = 10) and wildtype (n = 10) tumors. Results are shown in Extended Data Fig. \ref{fig:ex_data6}. For (1), we trained a ResNet50 model using conventional cross-entropy versus a weakly supervised patch-based contrastive learning, or patchcon. Both models were initialized using ImageNet pretrained weights \cite{Deng2009-hr} and trained for 10 epochs without additional hyperparameter tuning. For (2), the patchcon pretrained ResNet model from (1) was held fixed and we trained a single linear classification layer versus a transformer model with 3 multi-headed attention layers. Each model was trained for 10 epochs. For (3), only the transformer model was retrained using variable percentages of labels masked. We tested 0\%, 33\%, and 66\% of labels provided as input, which corresponded to 0, 1, and 2 labels provided for our dataset. Each model was trained using the same contrastive pre-trained ResNet SRH encoder to isolate the effect of label masked training on classifier performance. Results of ablation studies can be found in Extended Data Fig. \ref{fig:ex_data6}.

\subsection{Molecular heatmap generation}
Leveraging our previous work on semantic segmentation of SRH images \cite{Hollon2020-ez, Hollon2020-oj}, we densely sample patches at 100 pixel step size, which allows for local probability pooling from overlapping patch predictions. A major contribution of this work is the integration of our tumor segmentation model and DeepGlioma into a single interpretable heatmap for both molecular genetic and molecular subgroup predictions. The tumor segmented regions are retained and the normal/nondiagnostic regions are converted to grayscale in order to indicate these regions were not candidates for molecular prediction. Each molecular genetic heatmap is generated by averaging the output predictions from patches that overlap for any given pixel in the heatmap. Molecular subgroup heatmaps are more challenging and require integrating the molecular genetic predictions that are necessary for subgroup classification. To address this challenge, we use a molecular subgroup-specific conditional mask combined with IDH predictions to generate an interpretable and spatially consistent molecular subgroup heatmap. The most straightforward molecular subgroup heatmap is for glioblastoma, IDH-wildtype heatmap, generated as: 
\begin{equation}
p^\text{GBM}_{i,j} = 1 - p_\text{IDH}(\boldsymbol{x}_{i,j})
\end{equation}
such that $i, j$ corresponds to the whole slide height and width indices and $p_{IDH}(\boldsymbol{x}_{i,j})$ is the IDH prediction at the corresponding spatial location. In contrast, molecular oligodendrogliomas and astrocytomas require a conditional molecular mask to segment regions that meet specific molecular subgroup criteria. Molecular oligodendroglioma heatmaps are generated as:
\begin{equation}
p^\text{Oligo.}_{i,j} = \underbrace{\left[ p_\text{IDH}(\boldsymbol{x}_{i,j}) > \tau \land p_\text{1p19q}(\boldsymbol{x}_{i,j}) > \phi \right]}_\text{Conditional molecular mask} \cdot  p_\text{IDH}(\boldsymbol{x}_{i,j})
\end{equation}
with the binarized conditional molecular mask identifying heatmap regions that are above hyperparameter thresholds $\tau$ and $\phi$ for IDH and 1p19q codeletion, respectively. Molecular astrocytomas heatmaps are generated as:
\begin{equation}
p^\text{Astro.}_{i,j} = \underbrace{\left[p_{IDH}(\boldsymbol{x}_{i,j}) > \tau  \land \left[p_{1p19q}(\boldsymbol{x}_{i,j}) < \phi \lor p_\text{ATRX}(\boldsymbol{x}_{i,j}) > \pi \right]\right]}_\text{Conditional molecular mask} \cdot  p_\text{IDH}(\boldsymbol{x}_{i,j})
\end{equation}
where $\tau, \phi, \pi$ are all hyperparameter thresholds. All thresholds were set to 0.5 in our model without hyperparameter tuning to avoid overfitting. Conditional molecular masking encodes the spatial locations where the molecular subgroup conditions are instantiated and the IDH prediction provides the representative probability distribution for the molecular subgroup. Examples of molecular genetic and molecular subgroup heatmaps can be found in Extended Data Figs. \ref{fig:ex_data8} and \ref{fig:ex_data9}. Molecular heatmaps allowed for the evaluation of classification performance in different molecular settings. For example, DeepGlioma was able to correctly predict IDH-wildtype status in patients with recurrent mutations found in molecular glioblastomas, such as CDKNA1 and TERT promotor mutations (see Supplemental Fig. \ref{fig:sup_data1}).  Molecular heatmaps were also used to identify sources of DeepGlioma's classification errors. Potential sources, including low tumor infiltration and image quality, are presented in Supplemental Fig. \ref{fig:sup_data2}. Interactive web-based interface for DeepGlioma predictions can be found at \url{https://deepglioma.mlins.org}.

\subsection{Prospective multicenter clinical testing and sample size calculation}
We elected to perform prospective, international, multicenter clinical testing of DeepGlioma in order to adhere to the rigorous standards of responsible machine learning in healthcare \cite{Wiens2019-mh}. Our prospective clinical testing was designed using the same principles as a non-inferiority diagnostic clinical trial \cite{Hollon2020-oj}. NYU, UCSF, MUV, and UKK were all included as medical centers for prospective patient enrollment.

\paragraph{Primary testing endpoint} Our primary diagnostic endpoint was balanced classification accuracy ($\frac{\text{sensitivity}+\text{specificity}}{2}$) for diffuse glioma IDH mutational status. The control arm was conventional first-line laboratory molecular screening and the experimental arm was DeepGlioma predictions. IDH-1 immunohistochemistry (IHC) for somatic mutations at residue R132H is the most common first-line molecular diagnostic screening test. Dewitt et al. performed the largest and most clinically representative analysis of IDH mutation detection via IHC and sequencing methods, and determined that IDH1-R132H IHC has a balanced diagnostic accuracy of 91.4\% for adult-type diffuse gliomas (see Fig. \ref{fig:results}c for contingency tables) \cite{DeWitt2017-ju}. We used this value to set the expected accuracy for both the control and experimental arms, the equivalence limit was set to 10\%, power to 90\%, and alpha to 0.05\%, resulting in a sample size value of 135 patients. All sample size calculations were performed using the epiR package (version 2.0.46) in R (version 3.6.3). The majority of patients in our prospective cohort did not undergo both IHC and sequencing, therefore an accuracy value cannot be calculated for this group.

\paragraph{Secondary testing endpoint} Our secondary endpoint was to achieve improved classification performance compared to our previous methods for training deep computer vision models on SRH images for multiclass classification \cite{Hollon2020-ez, Hollon2020-oj}. End-to-end representation learning and classification can yield patch-based classification results that approach pathologist-level performance for histologic brain tumor classifcation. However, our early experiments on molecular classification indicated that contrastive pretraining and label embedding was advantageous for multi-label classification. Therefore, as a secondary endpoint, the control arm was established by training a ResNet50 model to classify the three mutually exclusive molecular subgroups using a conventional categorical cross-entropy loss function. This is equivalent to our previous model training method with the exception of different labels \cite{Hollon2020-ez, Hollon2020-oj}. Our experimental arm was DeepGlioma molecular subgroup predictions as shown in Algorithm \ref{algo1}. Secondary endpoint metric was overall multiclass classification accuracy (Fig. \ref{fig:results}d).

\subsection{Computational hardware and software}
All SRH images were processed using an Intel Core i76700K Skylake QuadCore 4.0 central processing unit (CPU) using our custom Python-based (version 3.8) mlins-package. We used the pydicom package (version 2.0.0) to process the SRH images from the NIO Imaging System. All archived postprocessed image patches were saved as 16-bit TIFF images and handled using the tifffile package (version 2021.1.14). All models were trained using the University of Michigan Advanced Research Computing (ARC) Armis2 high-performance computing cluster. Armis2 is a high-performance distributed computing environment that aligns with HIPAA privacy standards. Convolutional neural networks/visual encoders were trained on four NVIDIA Titan V100 graphical processing units (GPUs). Our genetic embedding model and classifiers were trained on eight NVIDIA 2080Ti GPUs. All custom code for training and inference can be found in our open-source DeepGlioma repository. Our models were implemented in PyTorch (version 1.9.0). We used the ImageNet pretrained ResNet50 model from torchvision (0.10.0). Scikit-learn (version 1.0.1) was used to compute performance metrics on model predictions at both training and inference. Additional dependencies and specifications can be found at our \href{https://github.com/MLNeurosurg/deepglioma}{MLNeurosurg/deepglioma} GitHub page.

\section{Data availability}
The genomic data for training the genetic embedding model are publicly available through the above-mentioned data repositories and all genomic data is provided in Extended Data Table 2. Institutional Review Board approval was obtained from all participating institutions for SRH imaging and data collection. Restrictions apply to the availability of raw patient imaging or genetic data, which were used with institutional permission through IRB approval for the current study, and are thus not publicly available. Please contact the corresponding authors (T.C.H. or D.A.O.) for any requests for data sharing. All requests will be evaluated based on institutional and departmental policies to determine whether the data requested is subject to intellectual property or patient privacy obligations. Data can only be shared for non-commercial academic purposes and will require a formal material transfer agreement.

\section{Code availability}
All code was implemented in Python (version 3.8) using PyTorch (1.9.0) as the primary machine learning framework. All code and scripts to reproduce the experiments of this paper are available on GitHub at \href{https://github.com/MLNeurosurg/deepglioma}{MLNeurosurg/deepglioma}. 

\section{Acknowledgments}
The results presented here are in whole or part based upon data generated by the TCGA Research Network: \url{https://www.cancer.gov/tcga}. We would like to thank Karen Eddy, Lin Wang, and Andrea Marshall for providing technical support and Tom Cichonski for editorial assistance.

\bibliographystyle{unsrt}
\bibliography{deepglioma.bib}

\clearpage\appendix

\section{Extended Data Figures}
\begin{enumerate}
    \item Overall workflow of intraoperative SRH and DeepGlioma.
    \item Training dataset.
    \item Multi-label contrastive learning for visual representation.
    \item SRH visual representation learning comparison.
    \item Diffuse glioma genetic embedding using global vectors.
    \item Ablation studies and cross-validation results.
    \item Patient demographic subgroup analysis of DeepGlioma IDH classification performance.
    \item DeepGlioma molecular subgroup analysis.
    \item Molecular genetic and molecular subgroup heatmaps.
    \item Evaluation of DeepGlioma on non-canonical diffuse gliomas.
\end{enumerate}

\section{Supplementary Data}
\subsection{Supplementary Data Tables (Available \href{https://doi.org/10.1038/s41591-023-02252-4}{Online})}
\begin{enumerate}
    \item DeepGlioma training dataset.
    \item Aggregated public diffuse glioma genomic dataset.
    \item Prospective multi-center testing dataset with DeepGlioma multi-label predictions.
    \item Prospective multi-center testing dataset with multiclass model predictions.
\end{enumerate}

\subsection{Supplementary Figures}
\begin{enumerate}
    \item Qualitative heatmap analysis of molecular glioblastomas.
    \item Error Analysis.
\end{enumerate}
\renewcommand{\figurename}{Extended Data Figure}
\setcounter{figure}{0}
\begin{figure*}[p!]
    \centering\includegraphics[width=\textwidth]{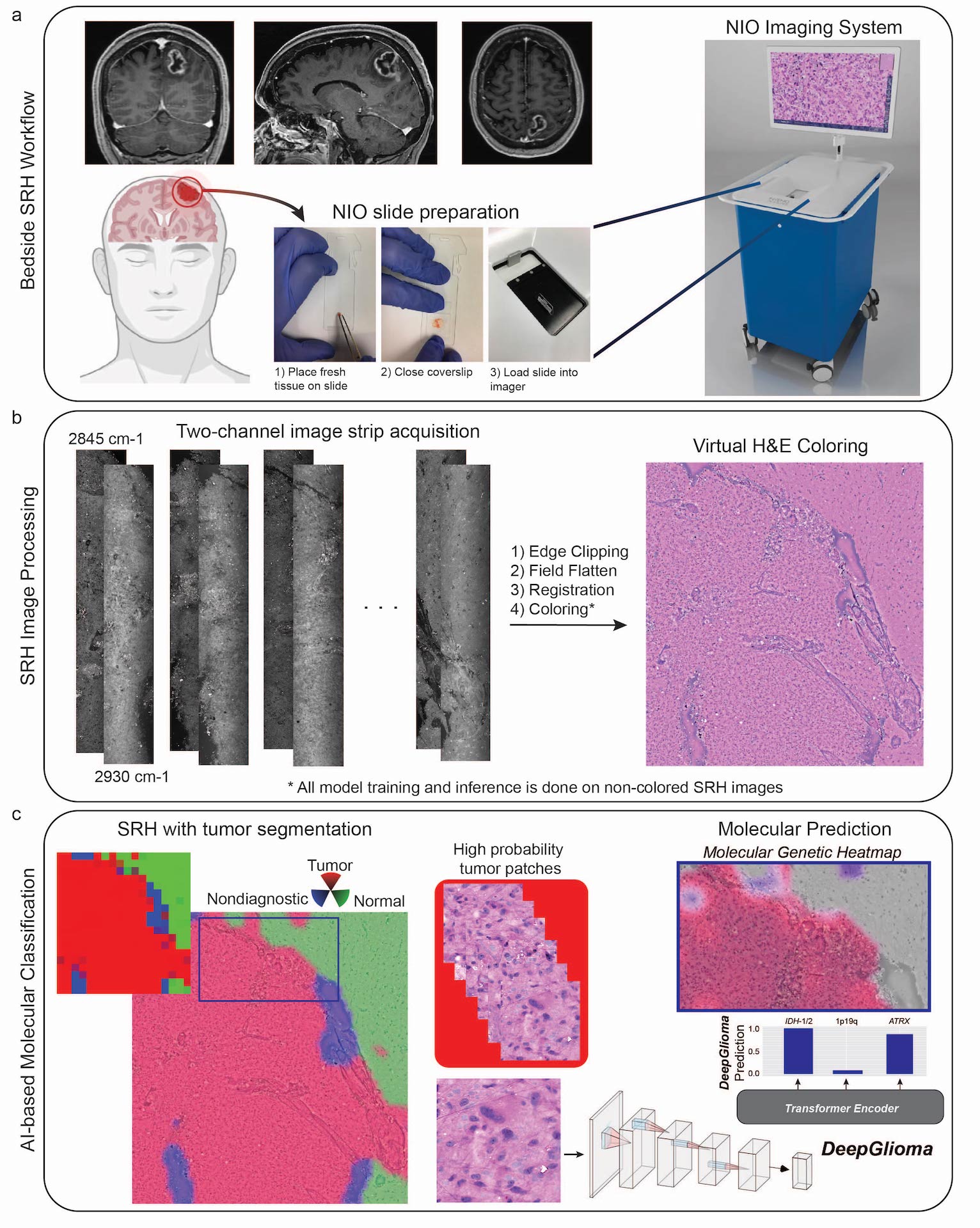}
    \caption{Overall workflow of intraoperative SRH and DeepGlioma.}
\end{figure*}
\addtocounter{figure}{-1}
\begin{figure*}[p!]\centering
  \caption{\textbf{Overall workflow of intraoperative SRH and DeepGlioma.} \textbf{a}, DeepGlioma for molecular prediction is intended for patients with clinical and radiographic evidence of a diffuse glioma who are undergoing surgery for tissue diagnosis and/or tumor resection. The surgical specimen is sampled from the patient’s tumor and directly loaded into a premade, disposable microscope slide with an attached coverslip. The specimen is loaded into the NIO Imaging System (Invenio Imaging, Inc., Santa Clara, CA) for rapid optical imaging. \textbf{b}, SRH images are acquired sequentially as strips at two Raman shifts, 2845 cm\textsuperscript{-1} and 2930 cm\textsuperscript{-1}. The size and number of strips to be acquired is set by the operator who defines the desired image size. Standard image sizes range from 1-5 mm\textsuperscript{2} and image acquisition time ranges from 30 seconds to 3 minutes. The strips are edge-clipped, field-flattened, and registered to generate whole slide SRH images, which are then used for both DeepGlioma training and inference. Additionally, whole slide images can be colored using a custom virtual H\&E color scheme for review by the surgeon or pathologist \cite{Orringer2017-nn}. \textbf{c}, For AI-based molecular diagnosis, the whole slide image is split into non-overlapping 300$\times$300-pixel patches and each patch undergoes a feedforward pass through a previously trained network to segment the patches into tumor regions, normal brain, and nondiagnostic regions \cite{Hollon2020-oj}. The tumor patches are then used by DeepGlioma at both training and inference to predict the molecular status of the patient’s tumor. }\label{fig:ex_data1}
\end{figure*}
\begin{figure*}[p!]
    \centering\includegraphics[width=\textwidth]{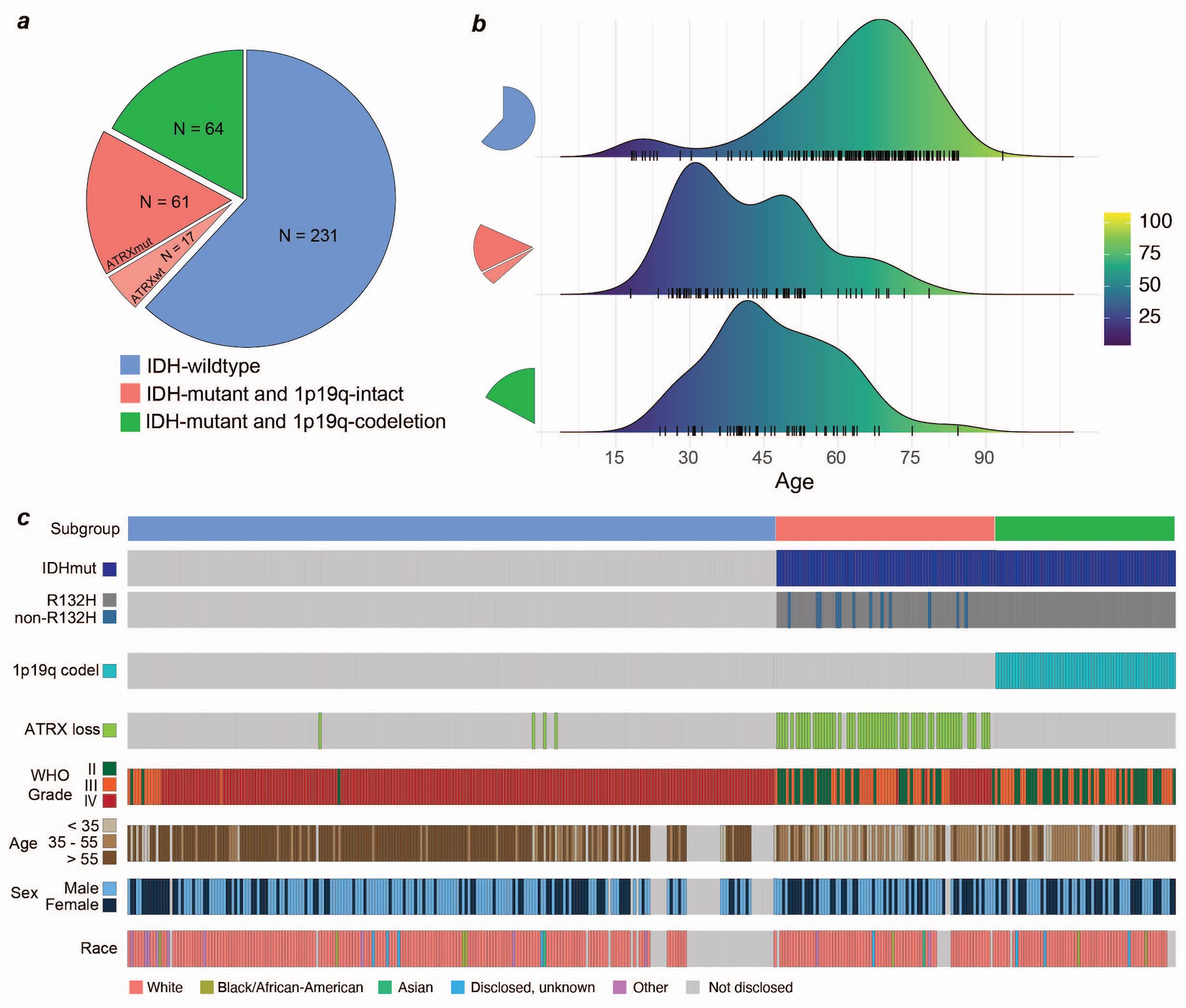}
    \caption{\textbf{Training dataset.} The UM adult-type diffuse gliomas dataset used for model training. The UM training set consisted of a total of 373 patients who underwent a biopsy or brain tumor resection. Dataset generation occurred over a 6-year period, from November 2015 through November 2021. \textbf{a}, The distribution of patients by molecular subgroup. IDH-wildtype gliomas consisted of 61.9\% (231/373) of the total dataset, IDH-mutant/1p19q-codeleted tumors consisted of 17.2\% (64/373), and IDH-mutant/1p19q-intact tumors consisted of 21.\% (78/373). Our dataset distribution of molecular subgroups is consistent with reported distributions in large-scale population studies. ATRX mutations were found in the majority of IDH-mutant/1p19q-intact patients (78\%), also concordant with previous studies \cite{Cancer_Genome_Atlas_Research_Network2015-ms}. \textbf{b}, The age distribution for each of the molecular subgroups is shown. The average age of IDH-wildtype patients was 62.6 $\pm$ 15.4 years and IDH-mutant patients was 44.6 $\pm$ 13.8 years. The average patient age of IDH-mutant/1p19q-codeleted group was 47.0 $\pm$ 12.9 years, and that of IDH-mutant/1p19-intact was 42.5 $\pm$ 14.1 years. \textbf{c}, Individualized patient characteristics and mutational status are shown by molecular subgroups. We report the WHO grade based on pathologic interpretation at the time of diagnosis. Because many of the patients were treated prior to the routine use of molecular status alone to determine WHO grade, several patients have IDH-wildtype lower grade gliomas (grade II or III) or IDH-mutant glioblastomas (grade IV). The discordance between histologic features and molecular features has been well documented \cite{Cancer_Genome_Atlas_Research_Network2015-ms} and is a major motivation for the present study.}
    \label{fig:ex_data2}
\end{figure*}
\begin{figure*}[p!]
    \centering\includegraphics[width=\textwidth]{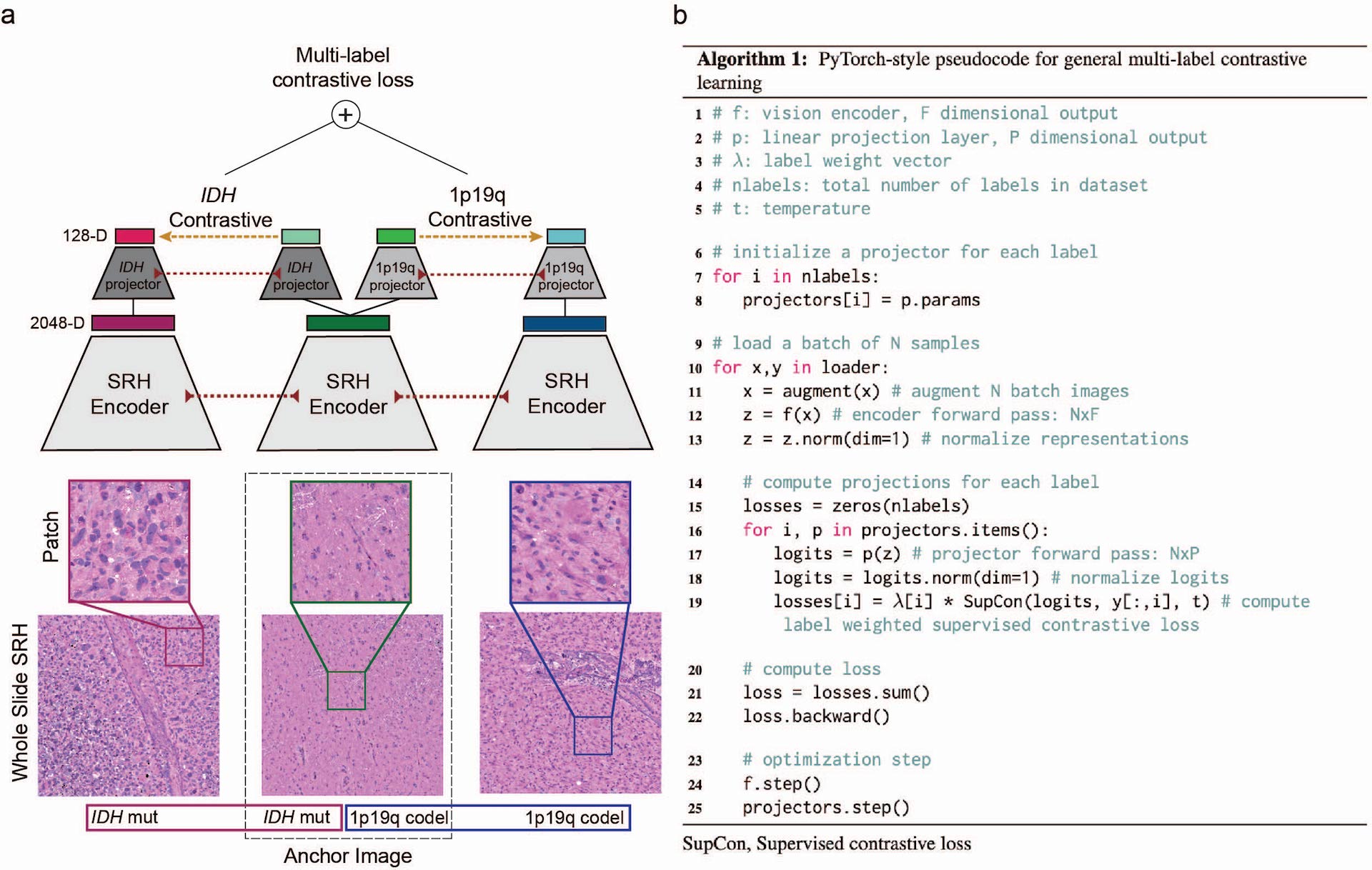}
    \caption{\textbf{Multi-label contrastive learning for visual representation.} Contrastive learning for visual representation is an active area of research in computer vision \cite{Chen2020-uz}. While the majority of research has focused on self-supervised learning, supervised contrastive loss functions have been underexplored and provide several advantages over supervised cross-entropy losses \cite{Khosla2020-hy}. Unfortunately, no straightforward extension of existing contrastive loss functions, such as InfoNCE and NT-Xent \cite{Chen2020-uz}, can accommodate  multi-label supervision. Here, we propose a simple and general extension of supervised contrastive learning for multi-label tasks and present the method in the context of patch-based image classification. \textbf{a}, Our multi-label contrastive learning framework starts with a randomly sampled anchor image with an associated set of labels. Within each minibatch a set of positive examples are defined for each label of the anchor image that shares the same label status. All images in the minibatch undergo a feedforward pass through the SRH encoder (red dotted lines indicate weight sharing). Each image representation vector (2048-D) is then passed through multiple label projectors (128-D) in order to compute a contrastive loss for each label (yellow dashed line). The scalar label-level contrastive loss is then summed and backpropagated through the projectors and image encoder. The multi-label contrastive loss is computed for all examples in each minibatch. \textbf{b}, PyTorch-style pseudocode for implementation of our proposed multi-label contrastive learning framework is shown. Note that this framework is general and can be applied to any multi-label classification task. We call our implementation \textit{patchcon} because individual image patches are sampled from whole slide SRH images to compute the contrastive loss. Because we use a single projection layer for each label and the same image encoder is used for all images, the computational complexity is linear in the number of labels.}\label{fig:ex_data3}
\end{figure*}

\clearpage
\begin{figure*}[p!]
    \centering
    \includegraphics[width=\textwidth]{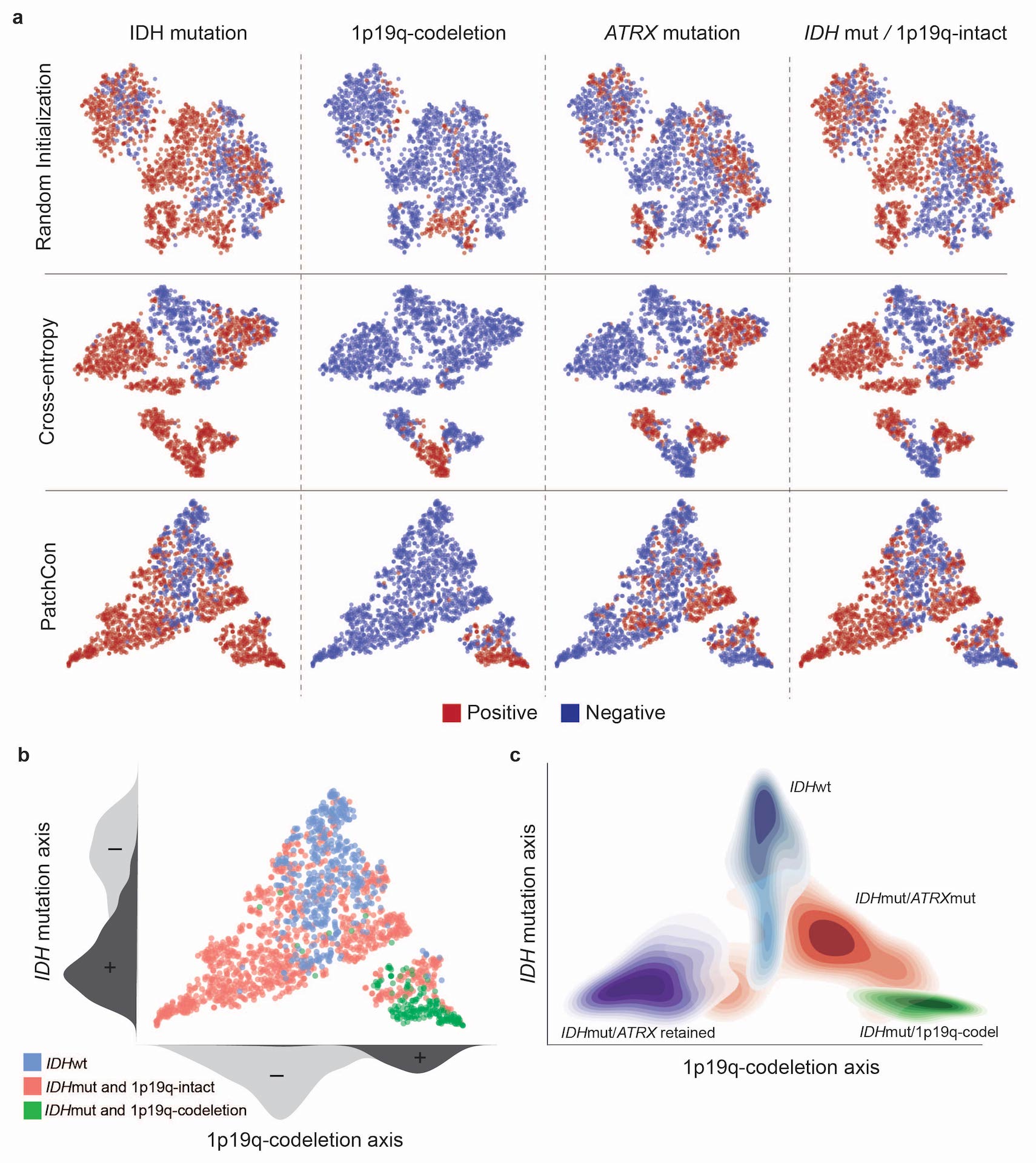}
    \caption{SRH visual representation learning comparison. }
    \label{fig:ex_data4}
\end{figure*}
\addtocounter{figure}{-1}
\begin{figure*}[p!]\centering
  \caption{\textbf{SRH visual representation learning comparison.} \textbf{a}, SRH patch representations of a held-out validation set are plotted. Patch representations from a ResNet50 encoder randomly initialized (top row), trained with cross-entropy (middle row), and PatchCon (bottom row) are shown. Each column shows binary labels for the listed molecular diagnostic mutation or subgroup. A randomly initialized encoder shows evidence of clustering because patches sampled from the same patient are correlated and can have similar image features. Training with a cross-entropy loss does enforce separability between some of the labels; however, there is no discernible low-dimensional manifold that disentangles the label information. Our proposed multi-label contrastive loss produced embeddings that are more uniformly distributed in representation space than cross-entropy. Uniformity of the learned embedding distribution is known to be a desirable feature of contrastive representation learning. \textbf{b}, Qualitative analysis of the SRH patch embeddings indicates that data are distributed along two major axes that correspond to IDH mutational status and 1p19q-codeletion status. This distribution produces a simplex with the three major molecular subgroups at each of the vertices. These qualitative results are reproduced in the prospective testing cohort shown in Figure \ref{fig:results}e. \textbf{c}, The contour density plots for each of the major molecular subgroups are shown to summarize the overall embedding structure. IDH-wildtype images cluster at the apex and IDH-mutant tumors cluster at the base. Patients with 1p19q-intact are closer to the origin and 1p19q-codeleted tumors are further from the origin.}
\end{figure*}

\clearpage
\begin{figure*}[p!]
    \centering
    \includegraphics[width=\textwidth]{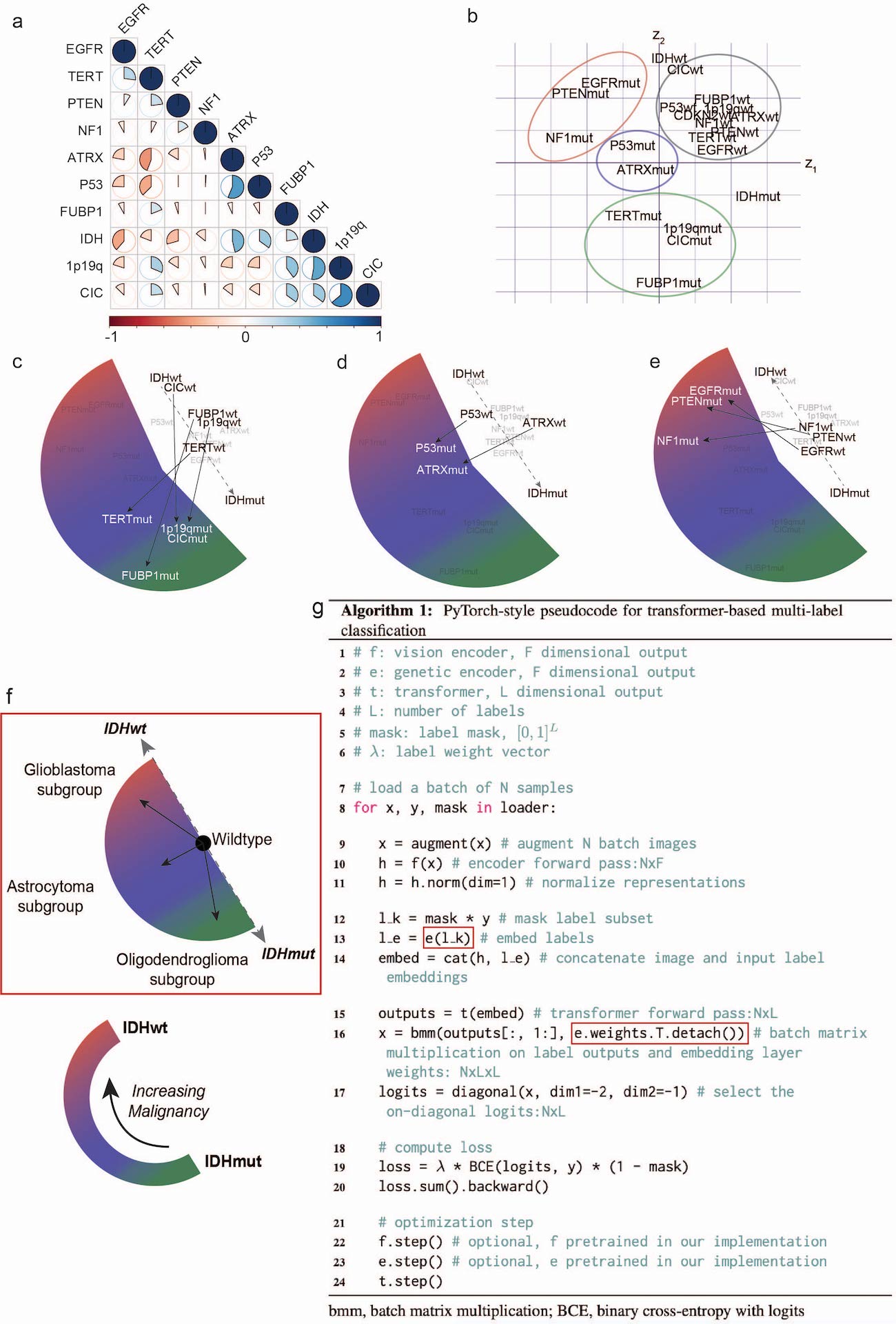}
    \caption{Diffuse glioma genetic embedding using global vectors.}
    \label{fig:ex_data5}
\end{figure*}
\addtocounter{figure}{-1}
\begin{figure*}[p!]
    \centering\caption{\textbf{Diffuse glioma genetic embedding using global vectors.} Embedding models transform discrete variables, such as words or gene mutational status, into continuous representations that populate a vector space such that location, direction, and distance are semantically meaningful. Our genetic embedding model was trained using data sourced from multiple public repositories of sequenced diffuse gliomas (Extended Data Table 2). We used a global vector embedding objective for training \cite{Pennington2014-dx}. \textbf{a}, A subset of the most common mutations in diffuse gliomas is shown in the co-occurrence matrix. Data were collected from multiple public repositories and aggregated to generate a single co-occurrence matrix for global vector embedding training. \textbf{b}, The learned genetic embedding vector space with the 11 most commonly mutated genes shown. Both the mutant and wildtype mutational statuses (N = 22) are included during training to encode the presence or absence of a mutation. Genes that co-occur in specific molecular subgroups cluster together within the vector space, such as mutations that occur in (\textbf{c}) IDH-mutant, 1p19q-codel oligodendrogliomas (green), (\textbf{d}) IDH-mutant, ATRX-mutant diffuse astrocytomas (blue), and (\textbf{e}) IDH-wildtype glioblastoma subtypes (red). Additionally, wildtype genes (black) form a single cluster with gene mutations organized in a radial pattern. Radial traversal of the embedding space defines clinically meaningful linear substructures \cite{Pennington2014-dx} corresponding to molecular subgroups. \textbf{f}, Corresponding to the known clinical and prognostic significance of IDH mutations in diffuse gliomas, IDH mutational status determines the axis along which increasing malignancy is defined in our genetic embedding space. \textbf{g}, PyTorch-style pseudocode for transformer-based masked multi-label classification. Inputs to our masked multi-label classification algorithm are listed in lines 1-5. The vision encoder and genetic encoder are pretrained in our implementation but can be randomly initialized and trained end-to-end. The label mask is an L-dimensional binary mask with a variable percentage of the labels removed and subsequently predicted in each feedforward pass. An image $x$ is augmented and undergoes a feedforward pass through the vision encoder $f$. The image representation is then $\ell^2$ normalized. The labels are embedded using our pretrained genetic embedding model and the label mask is applied. The label embeddings are then concatenated with the image embedding and passed into the transformer encoder as input tokens. Unlike previous transformer-based methods for multi-label classification \cite{Lanchantin2020-mz}, we enforce that the transformer encoder outputs into the same vector space as the pretrained genetic embedding model. We perform a batch matrix multiplication with the transformer outputs and the embedding layer weights. The main diagonal elements are the inner product between the transformer encoder output and the corresponding embedding weight values. We then compute the masked binary cross-entropy loss. In our implementation, this is used to train the transformer encoder model only.}
\end{figure*}
\clearpage
\clearpage
\begin{figure*}[p!]
    \centering
    \includegraphics[width=\textwidth]{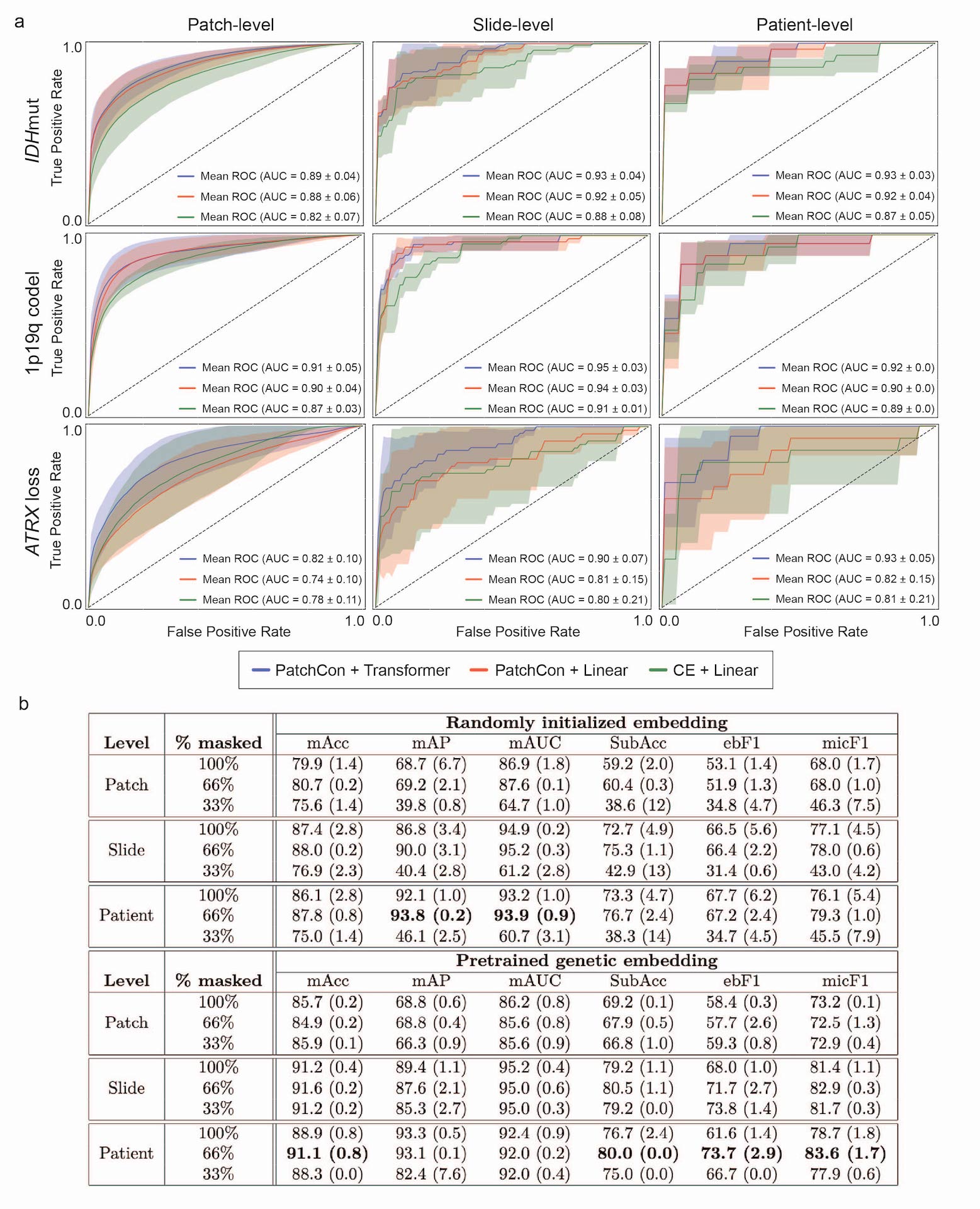}
    \caption{Ablation studies and cross-validation results.}
    \label{fig:ex_data6}
\end{figure*}
\clearpage
\addtocounter{figure}{-1}
\begin{figure*}[p!]
    \centering\caption{\textbf{Ablation studies and cross-validation results.} We conducted three main ablation studies to evaluate the following model architectural design choices and major training strategies: (1) cross-entropy versus contrastive loss for visual representation learning, (2) linear versus transformer-based multi-label classification, and (3) randomly initialized versus pretrained genetic embedding. \textbf{a}, The first two ablation studies are shown in the panel and the details of the cross-validation experiments are explained in the Methods section (see ‘Ablation Studies’). Firstly, a ResNet50 model was trained using either cross-entropy or patchcon. The patchcon trained image encoder was then fixed. A linear classifier and transformer classifier were then trained using the \textit{same} patchcon image encoder in order to evaluate the performance boost from using a transformer encoder. This ablation study design allows us to evaluate (1) and (2). The columns of the panel correspond to the three levels of prediction for SRH image classification: patch-, slide-, and patient-level. Each model was trained three times on randomly sampled validation sets and the average ($\pm$ standard deviation) ROC curves are shown for each model. Each row corresponds to the three molecular diagnostic mutations we aimed to predict using our DeepGlioma model. The results show that patchcon outperforms cross-entropy for visual representation learning and that the transformer classifier outperforms the linear classifier for multi-label classification. Note that the boost in performance of the transformer classifier over the linear model is due to the deep multi-headed attention mechanism learning conditional dependencies between labels in the context of specific SRH image features (i.e. not improved image feature learning because the SRH encoder weights are fixed). \textbf{b}, We then aimed to evaluate (3). Similar to the above, a single ResNet50 model was trained using patchcon and the encoder weights were fixed for the following ablation study to isolate the contribution of random initialization versus pretraining of the genetic embedding layer. Three mask label training regimes were tested and are presented in the tables: all input labels masked (100\%), two labels randomly masked (66\%), and one label randomly masked (33\%). The first row in the first table (100\% masked) is non-multimodal training, where no genetic information is provided at any point during training or inference. To better investigate the importance of masked label training, we report informative multi-label classification metrics. Genetic embedding pretraining outperformed random initialization on the majority of multi-label classification metrics. We found that 66\% input label masking, or randomly masking two of three diagnostic mutations, showed the best overall classification performance. We hypothesize that this results from allowing a single mutation to weakly define the genetic context while allowing supervision from the two masked labels to backpropagate through the transformer encoder. mAcc, mean label accuracy; mAP, mean average precision; mAUC, mean area under ROC curve; SubAcc, subset accuracy; ebF1, example-based F1 score; micF1, micro-F1 score.}
\end{figure*}
\clearpage
\begin{figure*}[p!]
    \centering
    \includegraphics[width=\textwidth]{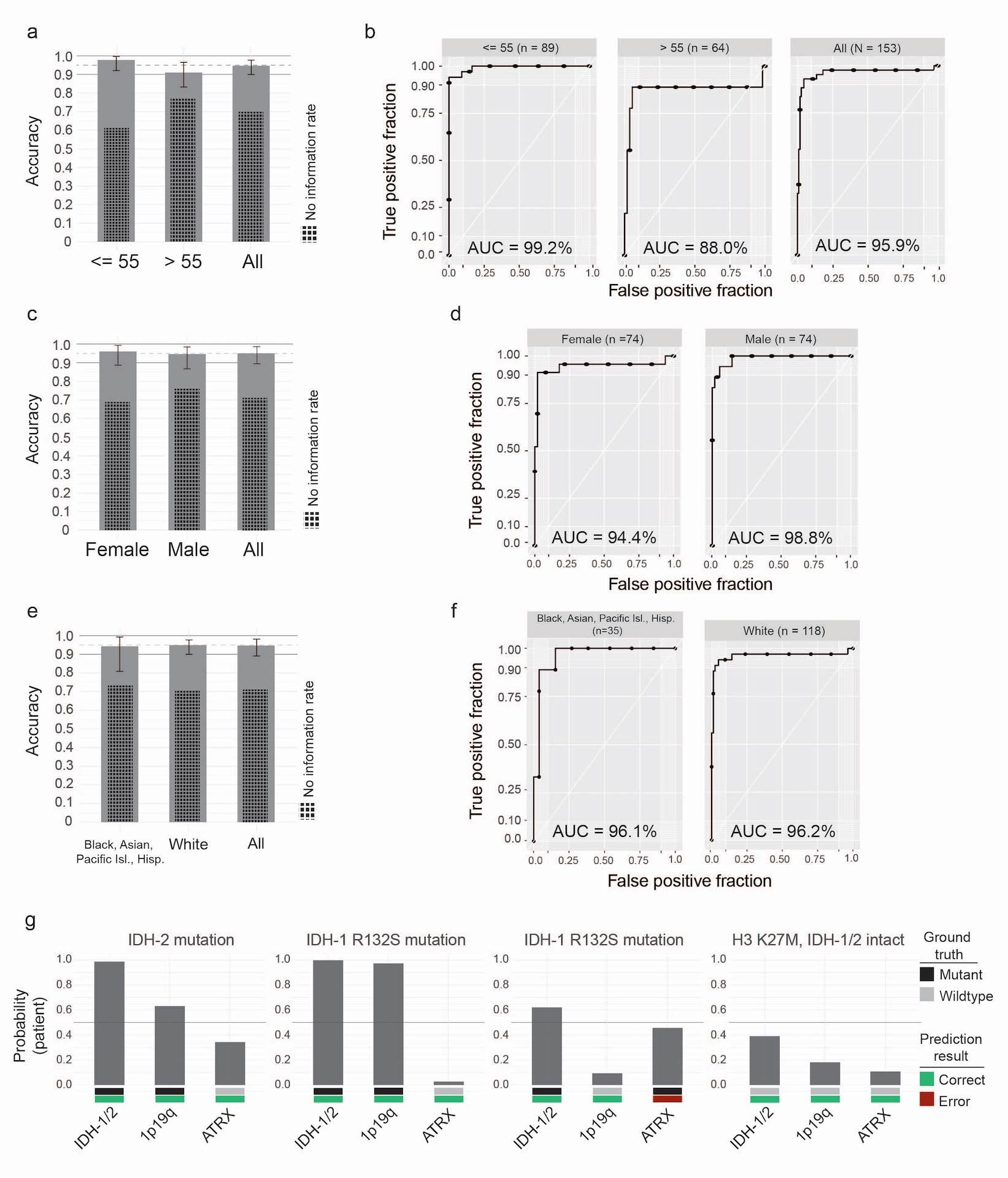}
    \caption{Patient demographic subgroup analysis of DeepGlioma IDH classification performance.}
    \label{fig:ex_data7}
\end{figure*}
\clearpage
\addtocounter{figure}{-1}
\begin{figure*}[p!]
    \centering\caption{\textbf{Patient demographic subgroup analysis of DeepGlioma IDH classification performance. }\textbf{a, b}, DeepGlioma performance for classifying IDH mutations stratified by patient age. Classification performance remains high in patients less than and greater than 55 years. IDH mutations are less common in patients greater than 55 years, causing class imbalance and resulting in a greater proportional drop in classification performance with false negative predictions. \textbf{c, d, } Classification performance stratified by sex, and \textbf{e, f} racial groups as defined by the National Insitute of Health (NIH). Classification performance remains high across all subgroup analyses. No information rate is the accuracy achieved by classifying all examples into the majority class. \textbf{g}, Subset of patients from the prospective cohort with non-canonical IDH mutations and a diffuse midline glioma, H3 K27M mutation. DeepGlioma correctly classified all non-canonical IDH mutations, including IDH-2 mutation. Moreover, DeepGlioma generalized to pediatric-type diffuse high-grade gliomas, including diffuse midline glioma, H3 K27-altered, in a zero-shot fashion as these tumors were not included in the UM training set. This patient was included in our prospective cohort because the patient was a 34-year-old adult at presentation.}
\end{figure*}
\clearpage
\begin{figure*}[p!]
    \centering
    \includegraphics[width=\textwidth]{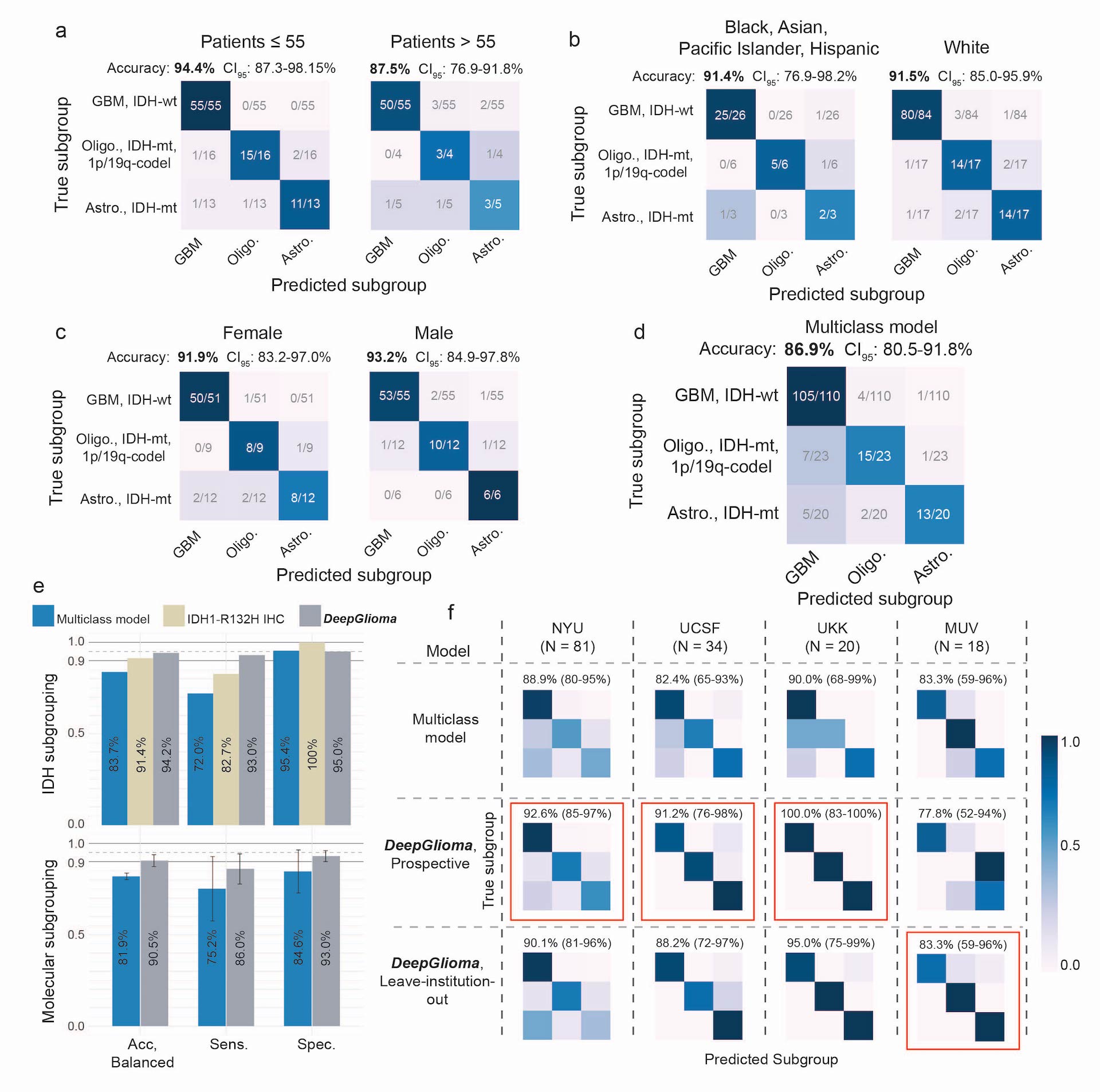}
    \caption{\textbf{DeepGlioma molecular subgroup analysis.} Multiclass classifcation performance for molecular subgroup prediction by DeepGlioma stratified by patient demographic information and prospective testing site is shown. Results stratified by (\textbf{a}) age, (\textbf{b}) race, and (\textbf{c}) sex are shown. Multiclass classification performance remained high in each patient demographic compared to the entire cohort. DeepGlioma was trained to generalize to all adult patients and to be agnostic to patient demographic information. \textbf{d}, Confusion matrix of our benchmark multiclass model trained using categorical cross-entropy. DeepGlioma outperformed the multiclass model by $+$4.6\% in overall patient-level diagnostic accuracy with a substantial improvement in differentiating molecular astrocytomas and oligodendrogliomas. \textbf{e}, Direct comparison of subgrouping performance for our benchmark multiclass model, IDH1-R132H IHC, and DeepGlioma. Performance metrics values are displayed. Molecular subgrouping mean and standard deviations are plotted for both IDH subgrouping and molecular subgrouping. These results provide evidence that multimodal training and multi-label prediction provide a performance boost over multi-class modeling. \textbf{f}, DeepGlioma molecular subgroup classification performance for each of the prospective testing medical centers is shown. Accuracy values with 95\% confidence intervals (in parentheses) are shown above the confusion matrices. Overall performance was stable across the three largest contributors of prospective patients. Performance on the MUV dataset was comparatively; however, some improvement was observed during the LIOCV experiments. Red indicates the best performance.}
    \label{fig:ex_data8}
\end{figure*}
\clearpage
\begin{figure*}[p!]
    \centering
    \includegraphics[width=\textwidth]{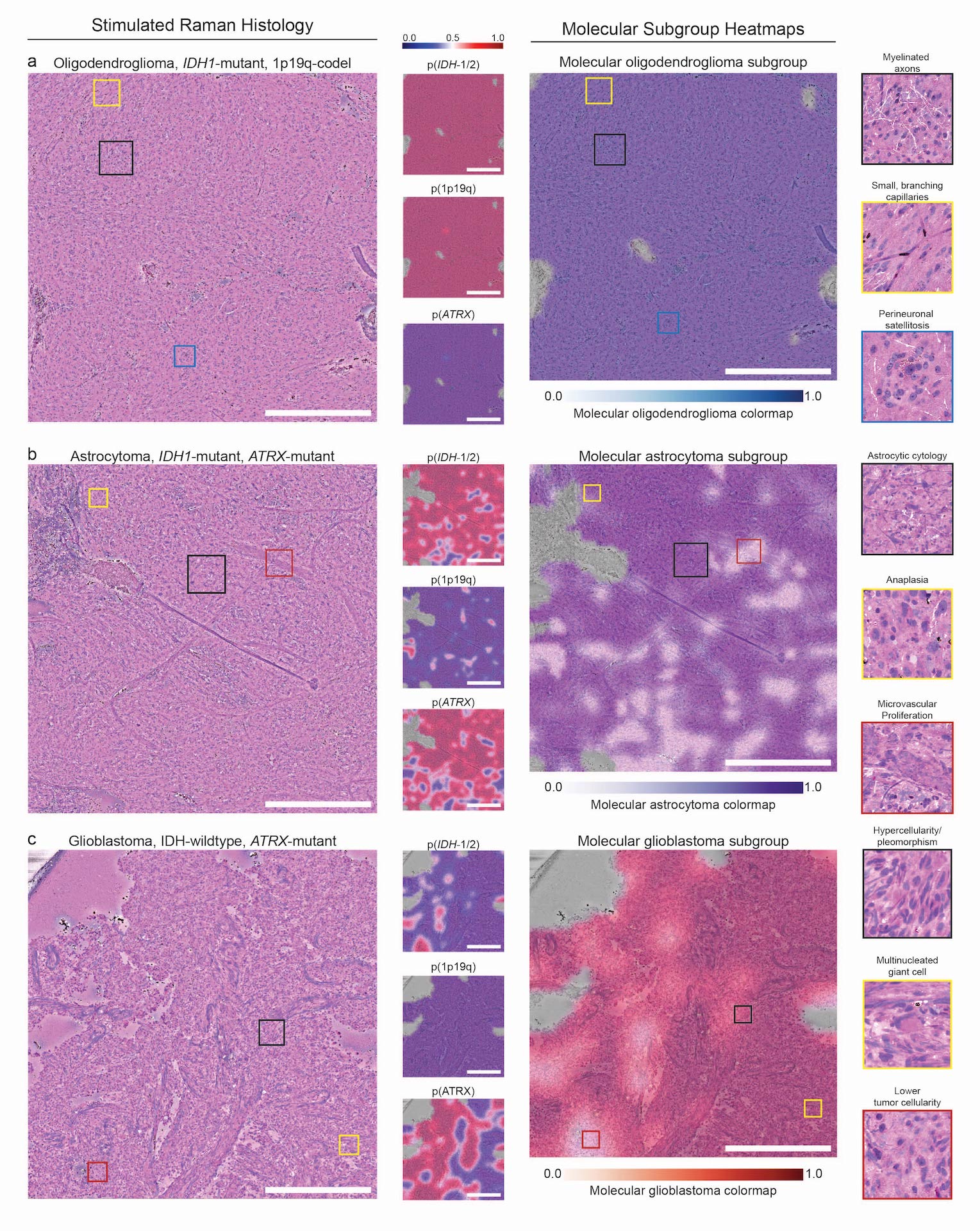}
    \caption{Molecular genetic and molecular subgroup heatmaps.}
    \label{fig:ex_data9}
\end{figure*}
\clearpage
\addtocounter{figure}{-1}
\begin{figure*}[p!]
    \centering\caption{\textbf{Molecular genetic and molecular subgroup heatmaps.} DeepGlioma predictions are presented as heatmaps from representative patients included in our prospective clinical testing dataset for each diffuse glioma molecular subgroup. \textbf{a}, SRH images from a patient  with a molecular oligodendroglioma, IDH-mutant, 1p19q-codel. Uniform high probability prediction for both IDH and 1p19q-codel and corresponding low ATRX mutation prediction. SRH images show classic oligodendroglioma features, including small, branching 'chicken-wire' capillaries and perineuronal satellitosis. Oligodendroglioma molecular subgroup heatmap shows expected high prediction probablity throughout the dense tumor regions. \textbf{b}, A molecular astrocytoma,  IDH-mutant, 1p19q-intact and ATRX-mutant is shown. Astrocytoma molecular subgroup heatmap shows some regions of lower probability that may be related to the presence of image features found in glioblastoma, such as microvascular proliferation. However, regions of dense hypercellularity and anaplasia are correctly classified as IDH mutant. These findings indicate DeepGlioma's IDH mutational status predictions are not determined solely by conventional cytologic or histomorphologic features that correlate with lower grade versus high grade diffuse gliomas. \textbf{c}, A glioblastoma, IDH-wildtype tumor is shown. Glioblastoma molecular subgroup heatmap shows high confidence throughout the tumor specimen. Additionally, this tumor was also ATRX mutated, which is known to occur in IDH-wildtype tumors \cite{Cancer_Genome_Atlas_Research_Network2015-ms}. Despite the high co-occurence of IDH mutations with ATRX mutations, DeepGlioma was able to identify image features predictive of ATRX mutations in a molecular glioblastoma. Because ATRX mutations are not diagnostic of molecular glioblastomas, the ATRX prediction does not affect the molecular subgroup heatmap (see `Molecular heatmap generation' section in Methods). Additional SRH images and DeepGlioma prediction heatmaps can be found at our interactive web-based viewer \url{https://deepglioma.mlins.org}.}
\end{figure*}
\clearpage
\begin{figure*}[p!]
    \centering
    \includegraphics[width=\textwidth]{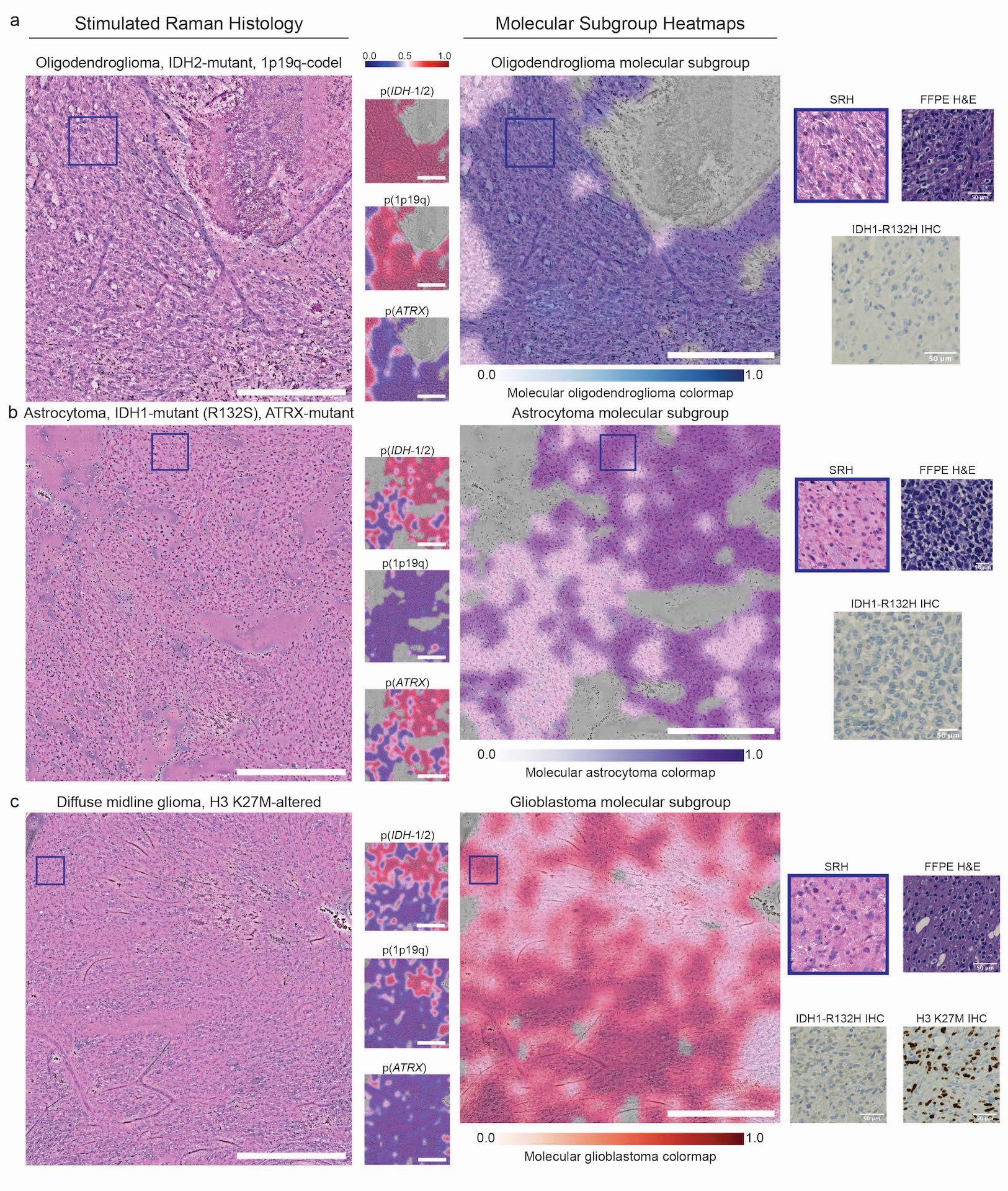}
    \caption{Evaluation of DeepGlioma on non-canonical diffuse gliomas.}
    \label{fig:ex_data10}
\end{figure*}
\clearpage
\addtocounter{figure}{-1}
\begin{figure*}[p!]
    \centering\caption{\textbf{Evaluation of DeepGlioma on non-canonical diffuse gliomas.} A major advantage of DeepGlioma over conventional immunohistochemical laboratory techniques is that it is not reliant on specific antigens for effective molecular screening. \textbf{a}, A molecular oligodendroglioma with an IDH2 mutation is shown. DeepGlioma correctly predicted the presence of both an IDH mutation and 1p19q-codeletion. IDH1-R132H IHC performed on the imaged specimen is negative. The patient was younger than 55 and, therefore, required genetic sequencing in order to complete full molecular diagnostic testing using our current laboratory methods. \textbf{b}, A molecular astrocytoma with IDH1-R132S and ATRX mutations. DeepGlioma correctly identifies both mutations. \textbf{c}, A patient with a suspected adult-type diffuse glioma met inclusion criteria for the prospective clinical testing set. The patient was later diagnosed with a diffuse midline glioma, H3 K27-altered. DeepGlioma correctly predicted the patient to be IDH-wildtype without previous training on diffuse midline gliomas or other pediatric-type diffuse gliomas. We hypothesize that DeepGlioma can perform well on other glial neoplasms in a similar zero-shot fashion.}
\end{figure*}
\clearpage
\renewcommand{\figurename}{Supplemental Figure}
\setcounter{figure}{0}
\clearpage
\begin{figure*}[p!]
    \centering
    \includegraphics[width=\textwidth]{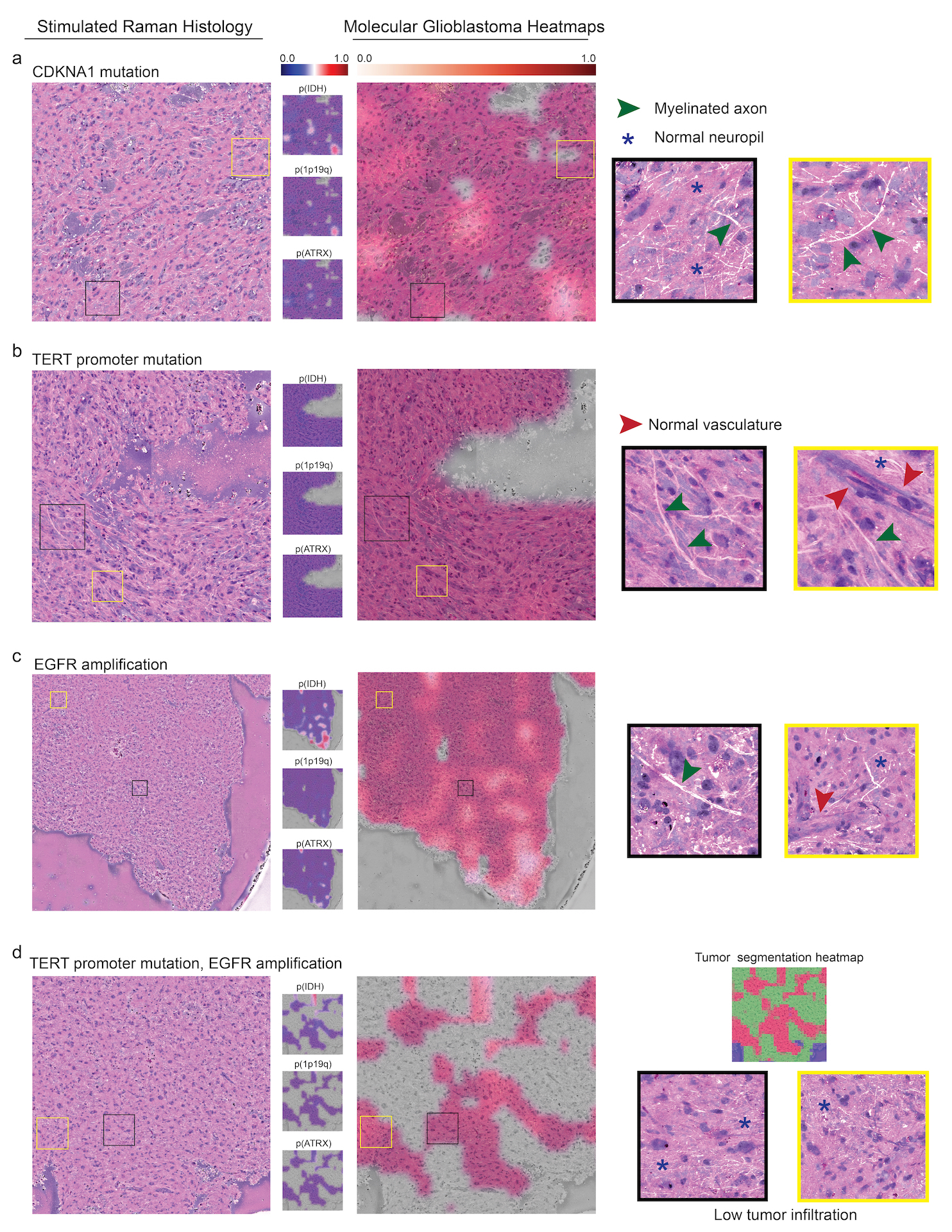}
    \caption{Qualitative heatmap analysis of molecular glioblastomas}
    \label{fig:sup_data1}
\end{figure*}
\clearpage
\addtocounter{figure}{-1}
\begin{figure*}[p!]
    \centering\caption{\textbf{Qualitative heatmap analysis of molecular glioblastomas} Molecular genetic and molecular glioblastoma subgroup heatmaps are shown for tumors that have recurrent mutations found in IDH-wildtype glioblastomas. Patients with (\textbf{a}) CDKNA1 mutation, (\textbf{b}) TERT promoter mutation, (\textbf{c}) EGFR amplification, and (\textbf{d}) both are shown. These cases demonstrate that in the absence of traditional high-grade glioma histologic features, such as microvascular proliferation or necrosis, DeepGlioma is able to correctly classify the molecular status of IDH-wildtype glioblastomas. High-magnification views of SRH demonstrate that in the setting of normal brain tissue infiltration, evidenced by healthy myelinated axons, neuropil, and normal vasculature, DeepGlioma predicts IDH-wildtype status for each of the molecular glioblastomas. \textbf{d}, DeepGlioma can correctly predict IDH-wildtype status in regions of low tumor infiltration, such that large regions of the specimen are classified as normal brain by the tumor segmentation model.}
\end{figure*}
\clearpage
\begin{figure*}[p!]
    \centering
    \includegraphics[width=\textwidth]{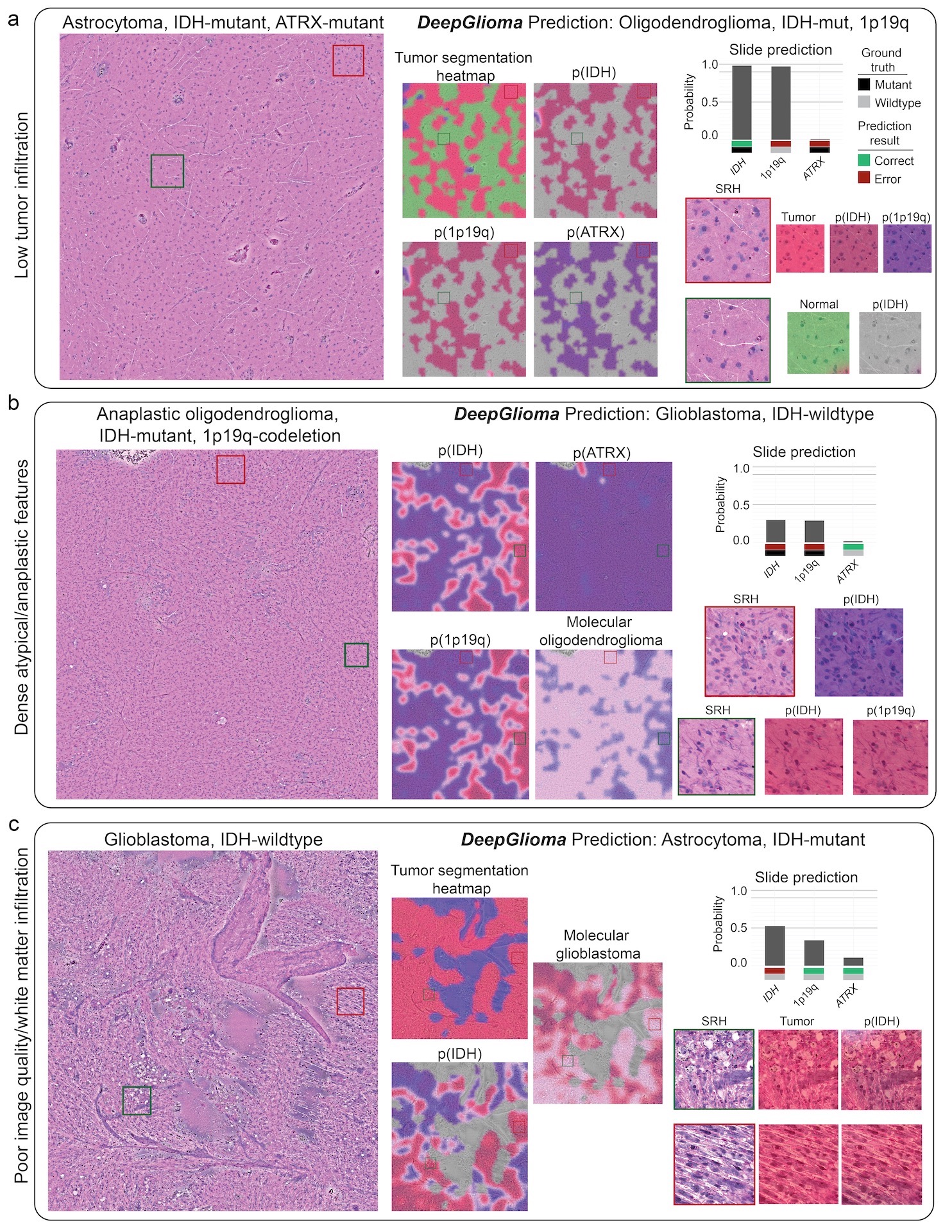}
    \caption{Error Analysis.}
    \label{fig:sup_data2}
\end{figure*}
\clearpage
\addtocounter{figure}{-1}
\begin{figure*}[p!]
    \centering\caption{\textbf{Error Analysis.} To better understand the source of DeepGlioma's classification errors, we performed an error analysis to identify some potential failure modes. \textbf{a}, Low tumor infiltration can result from at least two scenarios, intrinsically low tumor density within tumor core specimens or specimen sampling at non-core tumor margins. Differentiating these two states is challenging for any diagnostic system because it depends on both intrinsic tumor biology and accurate tissue sampling. DeepGlioma is unable to differentiate these two scenarios, but future work directed at classifying core versus non-core specimens is underway. Shown is an astrocytoma, IDH-mutant that was correctly classified as an IDH-mutant tumor, but incorrectly classified as having an 1p19q-codeletion. We hypothesize that this is due to lower tumor infiltration (red inset) often found in molecular oligodendrogliomas. The tumor segmentation heatmap shows that large regions of the specimens were classified as normal brain (green inset). \textbf{b}, Dense atypical or anaplastic features in IDH-mutant tumors remain a challenge is some instances. Shown is a patient with an anaplastic molecular oligodendroglioma with dense atypical (red inset) and astrocytic (green inset) image features. While large regions within the image were correctly classified as IDH-mutant and 1p19q-codeletion, whole slide inference produced an incorrect diagnosis. Despite dense astrocytic features throughout the specimen, large regions were correctly classified as 1p19q-codeleted. We believe that these errors can be ameliorated with additional training data and trainable whole slide inference methods. \textbf{c}, Poor imaging quality, often a result of Raman signal scattering due to white matter infiltration (red inset) \cite{Freudiger2008-gj}, may result in classification errors. Shown is an image from an infiltrative glioblastoma, IDH-wildtype. Rare regions of tissue injury (green inset) due to SRH imaging are artifactual and may lead to misdiagnosis. We believe these errors can be eliminated in future model versions by improved region-based or whole slide image filtering that explicitly evaluates image quality prior to DeepGlioma inference.}
\end{figure*}
\end{document}